\documentclass[]{ecai} 

\usepackage{latexsym}
\usepackage{amssymb}
\usepackage{amsmath}
\usepackage{amsthm}
\usepackage{booktabs}
\usepackage{enumitem}
\usepackage{graphicx}
\usepackage{color}
\usepackage{footmisc}
\usepackage{fancyhdr}

\usepackage{xcolor}
\usepackage{makecell}
\usepackage{arydshln}

\newcommand{\BibTeX}{B\kern-.05em{\sc i\kern-.025em b}\kern-.08em\TeX}

\begin{document}


\begin{frontmatter}


\paperid{2207} 


\title{Natural Mitigation of Catastrophic Interference: Continual Learning in Power-Law Learning Environments}


\newcommand{\gtlogo}{\raisebox{3.4pt}{\includegraphics[scale=0.04]{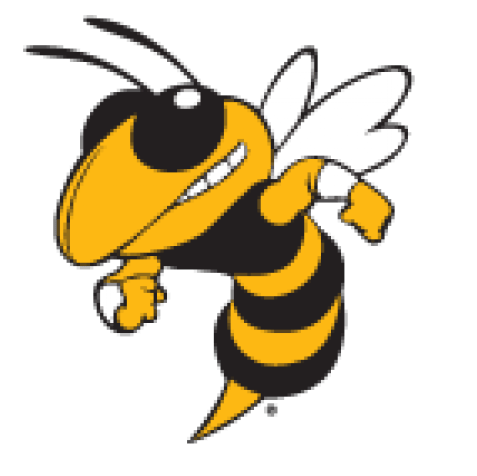}}}
\makeatletter
\def\thanks#1{\protected@xdef\@thanks{\@thanks
        \protect\footnotetext{#1}}}
\makeatother

\author[a]{Atith Gandhi \footnote{Equal contribution.\\ Supplementary materials - https://arxiv.org/abs/2401.10393}}
\author[a]{Raj Sanjay Shah \footnotemark}
\author[a]{Vijay Marupudi }
\author[a]{\fnms{Sashank}~\snm{Varma }\thanks{Corresponding author email - varma@gatech.edu}}


    
\address[a]{Georgia Institute of Technology\gtlogo}


\begin{abstract}

Neural networks often suffer from catastrophic interference (CI): performance on previously learned tasks drops off significantly when learning a new task. This contrasts strongly with humans, who can continually learn new tasks without appreciably forgetting previous tasks. Prior work has explored various techniques for mitigating CI and promoting continual learning such as regularization, rehearsal, generative replay, and context-specific components. This paper takes a different approach, one guided by cognitive science research showing that in naturalistic environments, the probability of encountering a task decreases as a power-law of the time since it was last performed. We argue that techniques for mitigating CI should be compared against the intrinsic mitigation in \emph{simulated naturalistic learning environments}. Thus, we evaluate the extent of the natural mitigation of CI when training models in \emph{power-law environments}, similar to those humans face. Our results show that natural rehearsal environments are better at mitigating CI than existing methods, calling for the need for better evaluation processes. The benefits of this environment include simplicity, rehearsal that is agnostic to both tasks and models, and the lack of a need for extra neural circuitry. In addition, we explore popular mitigation techniques in power-law environments to create new baselines for continual learning research.
\end{abstract}

\end{frontmatter}


\section{Introduction}
\label{introduction}

Humans learn to perform new skills throughout their lifetime without appreciable forgetting of old skills. Within the context of machine learning (ML) algorithms, the ability to incrementally acquire new knowledge while retaining previously learned experiences is known as \emph{continual learning} or \emph{lifelong learning} \cite{mitchell_etal, PARISI201954}. ML models are typically trained to perform a single task. They would be more useful if they could learn new tasks sequentially over time while remembering multiple old tasks. However, when models learn new tasks, this often leads to a drastic drop in the performance of old (i.e., previously learned) tasks. This phenomenon is known as \emph{catastrophic interference} (CI), and it is a major challenge for ML models. The current study looks to cognitive science to emulate the distribution of data observed in human-like learning environments and explores the \emph{natural mitigation of CI} in neural networks trained in such simulated environments.

Neural network models are capable of learning to perform a variety of tasks. However, when a new task is introduced, these models must be re-trained on the new task as well as all old tasks due to CI. Unfortunately, this is computationally inefficient. Ideally, neural networks should learn continuously, incrementally acquiring new tasks while minimally forgetting prior tasks. There is extensive research on methods to mitigate CI in ML models. These include weight regularization, selective forgetting, and memory replay to maintain the performance of previously learned tasks \cite{wang2023comprehensive}. These methods are typically tested on the availability of uniformly
distributed samples of each task, which is not representative of the regularity with which tasks repeat in natural environments. Here, we introduce a different approach, one that derives from cognitive science studies of naturalistic learning environments.

\subsection{Naturalistic cognitive learning environments}

\citet{doi:10.1111/j.1467-9280.1991.tb00174.x} found that human \emph{memory} is optimized for the natural distributions of environmental events such as the words in newspaper headlines, the utterances of parents around children learning to speak, and email messages. A Bayesian analysis revealed that the probability of needing to retrieve a particular item from memory declines as a power function of the time since that item was last retrieved. Their findings have been replicated \cite{Schooler1997} and extended to environments ranging from laboratory studies of human memory \cite{Anderson1997} to field studies of the social environment of chimpanzees \cite{Stevens2016}. Moreover, human \emph{learning} appears to be sensitive to this distribution, as human performance with practice at skills improves according to a power-law \cite{newell2013mechanisms}. Some studies have also proposed alternatives to this power-law of practice. \citet{Heathcote2000} argue that the exponential-law is a better fit than the power-law for data sources where individual samples cannot be aggregated, for example, as a practice function for individual learners. Given this competing account, we also consider the exponential naturalistic learning environment in this study. 

Moving beyond cognitive science, numerous real-world environments also follow naturalistic power-law distributions \footnote{See the supplementary material for a non-exhaustive list of environments where such frequency distributions are observed}. \textbf{We, therefore, argue that techniques for mitigating CI should be compared against potentially intrinsic mitigation in naturalistic learning environments where rehearsal follows the same power-law distribution.} This mimics learning systems (i.e., people) in the real world, who may not be storing examples of previously seen tasks in memory but rather ``rehearsing'' those examples that reoccur naturally.

\begin{figure*}[ht]
\centering
\includegraphics[width=0.9\textwidth]{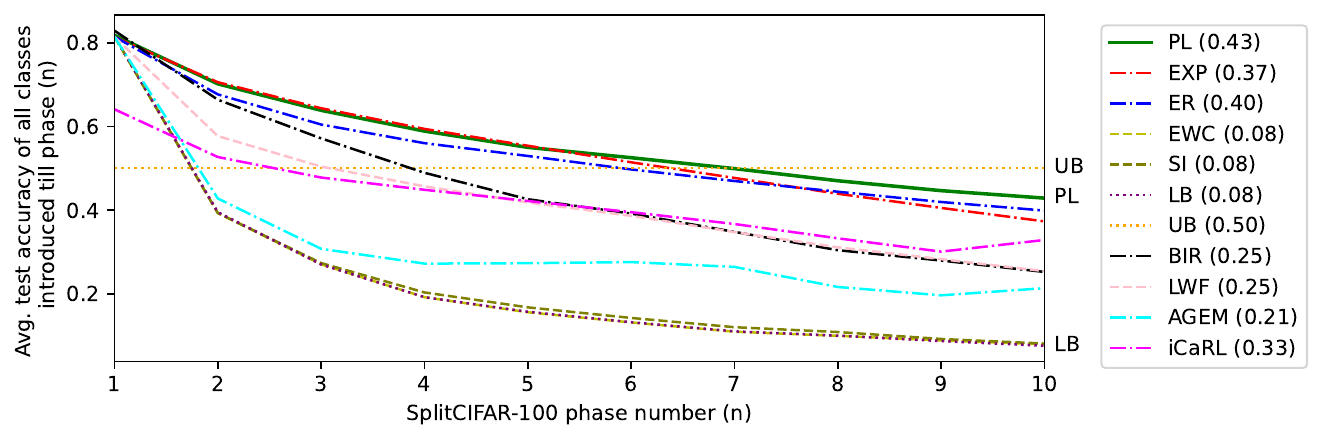}
\includegraphics[width=0.9\textwidth]{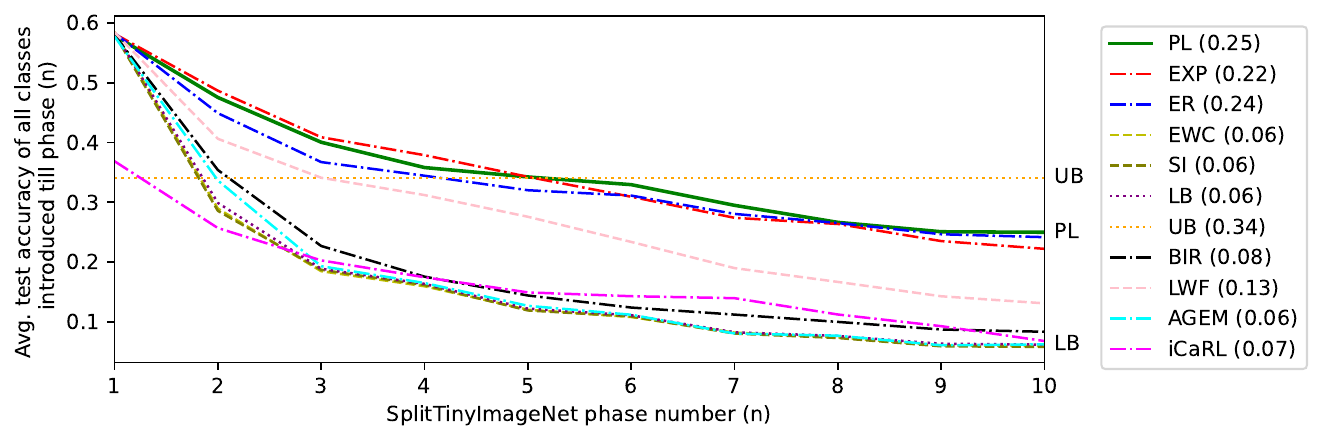}
\caption{Comparison of all the baselines and training environments on the SplitCIFAR-100 (above) and SplitTinyImageNet (below) class incremental learning scenarios with 10 phases. The values in the bracket indicate average test accuracy at the end of 10 phases. Note that EWC performs as poorly as the lower baseline across all phases for both datasets. LB: Lower Baseline, UB: Upper Baseline, EWC: Elastic Weight Consolidation, SI: Synaptic Intelligence, LwF: Learning without Forgetting, ER: Experience Replay, BIR: Brain-Inspired Replay, A-GEM: Averaged Gradient Episodic Memory, \textbf{PL: Power-law}, \textbf{Exp: Exponential}, iCaRL: Incremental Classifier and Representation Learning.}
\label{fig:paper_fig_1}
\end{figure*}

Taking inspiration from this characterization of natural environments \cite{doi:10.1111/j.1467-9280.1991.tb00174.x},  \citeauthor{lyndgaard2022catastrophic} (\citeyear{lyndgaard2022catastrophic}) investigated whether training in power-law environments mitigates CI. They trained a neural network on a sequence of tasks, with the proportion of training samples for a previous task decreasing as a power function of the number of intervening tasks. The results suggested that a power-law training environment might mitigate CI. However, their study included simplistic training tasks from the cognitive science literature i.e., sequentially learning the boolean functions AND, OR, XOR, and NAND. In particular, they did not use the more complex SplitMNIST, SplitCIFAR-100, or SplitTinyImageNet tasks that are standard in ML studies of continual learning. Furthermore, they did not compare training in power-law environments to the performance of other CI mitigation methods or an upper baseline. Here, we expand upon their work by conducting a comprehensive evaluation of the CI mitigation properties of power-law environments.


\subsection{Research questions}

The current study simulates a naturalistic power-law learning environment and assesses the extent of natural mitigation of CI. We evaluate it on standard problems in the continual learning literature -- SplitMNIST, SplitCIFAR-100, and SplitTinyImageNet -- and compare its performance to appropriate upper and lower baselines. Additionally, we compare its performance to representative CI mitigation approaches and also to training in other, non-power-law training environments. In more detail, the study addresses the following research questions:
\begin{enumerate}

    \item \textbf{Simulation:} Can we simulate rehearsal methodologies (i.e., power-law training environments) inspired by the naturalistic learning environments of humans? 
    \item \textbf{Natural mitigation:} What is the extent of CI mitigation when using the natural rehearsal of power-law training environments?
    \begin{enumerate}
        \item How does this training environment compare to current baselines in traditional training environments?
        \item How does this training environment compare to an alternative naturalistic environment (i.e., an exponential environment)?
        \item Does this natural mitigation of CI persist when the number of phases (i.e., tasks) increases?
        \end{enumerate}
    \item \textbf{Combination with other approaches:} In a power-law rehearsal environment, how well do prevalent approaches to mitigating CI perform, i.e., is their value in combining methods?
    

\end{enumerate}

\section{Related Work}

There is a considerable body of research in continual learning or lifelong learning in neural networks \cite{thrun_etal, MCCLOSKEY1989109, mitchell_etal, PARISI201954}.
Methods for mitigating CI largely fall into four major categories: context-specific components, parameter-isolation, regularization-based, and rehearsal-based methods. Some methods like Learning without Forgetting (LwF) \cite{li2017learning} fall into multiple categories. We review representative methods in each of the categories below.

Methods based on context-specific components use specific adapters or additional components for each task or context \cite{zhang2022continual, wang2023comprehensive}. These adapters need additional information about the contexts or tasks and incrementally need more computation units. A limitation of this approach is that this information and additional resources may not be readily available during the incremental addition of new tasks.

Parameter-isolation approaches \cite{mallya2018packnet, serra2018overcoming} seek to isolate neural network weights that have a greater probability of being relevant to previous tasks. \citeauthor{mallya2018packnet} attempt to identify and iteratively prune redundant parameters in a neural network to allow it to learn new tasks. Taking a different approach, Hard Attention Networks \cite{serra2018overcoming} mask different areas of neural network layers to explicitly allocate network capacity to different tasks. A limitation of this approach is that they are model architecture-dependent and may not generalize to all training algorithms.

Many recent works explore parameter regularization techniques \cite{kirkpatrick2017overcoming, zenke2017continual, li2017learning, rusu2022progressive}. One popular technique, Elastic Weight Consolidation (EWC) \cite{kirkpatrick2017overcoming, schwarz2018progress}, slows down adjustments to existing weights based on their importance to previously learned tasks. Similarly, \citeauthor{zenke2017continual} propose the use of \emph{intelligent synapses} that mimic the complexity of biological synapses. During training, the importance of each synapse is computed by considering its local contribution to the change in the global loss. This helps prevent the loss of information from changing synapses important for a particular old task while learning a new task. Many other regularization techniques attempt to determine the importance of weights during regularization \cite{lee2017overcoming} and increase the penalty for altering important weights. However, in the general case, neural network behavior is dependent on initialization, training order, and many other factors. This makes it difficult to accurately estimate the importance of weights to tasks.

Functional regularization methods discourage changes in outputs for a set of anchor points from previous rounds. These techniques look at regularization in the function space as opposed to the parameter space. For example, Learning without Forgetting \cite{li2017learning}, or LwF, performs knowledge distillation by keeping track of model outputs for the same input samples over different phases and computes a combined loss over the ground truth new task output with old model output stored in memory. A limitation of this technique is that it can potentially lead to error propagation and gradual erroneous behavior build-up over learning phases. Note that recent literature \cite{van2022three} shows that regularization methods fail to scale up with the complexity of tasks while rehearsal-based methods show better mitigation of CI.

Two of the methods discussed above, weight regularization and functional regularization, are orthogonal to the naturalistic rehearsal environments, and therefore the approaches can be combined. Section \ref{sec:combination} explores the mitigation of CI when implementing these methods in power-law training environments.

\subsection*{Rehearsal-based methods }
\begin{figure*}[htp]
\centering
\includegraphics[width=\textwidth]{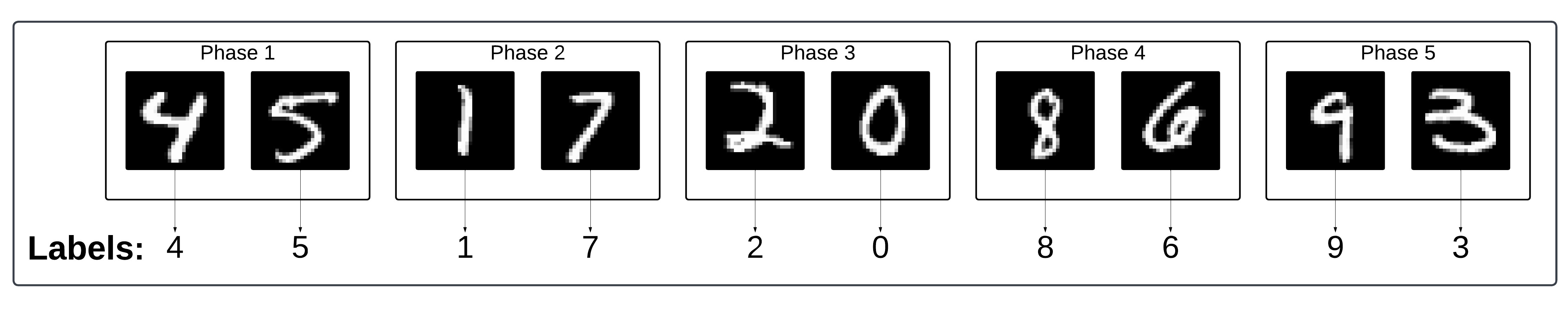}
\vspace{-7mm}
\caption{Example of the Class Incremental Learning scenario (SplitMNIST dataset): In each phase, new classes are incrementally added to the training.}
\label{fig:paper_class_cl_example}
\end{figure*}
The fourth category is composed of \textit{rehearsal-based methods}, also known as \textit{replay-based methods \footnote{We use the terms "rehearsal-based methods" and "replay-based methods" inter-changeably.}} \cite{deepSLDA, gem, onlineIncrementalLearningVector}. These methods use some learning samples from previous tasks as training samples upon the introduction of a new task. The samples of previous tasks can either be retained in a buffer upon introduction (iCaRL, ER) or generated by a separate generative model (DGR, BI-R). One prevalent rehearsal method, Experience Replay (ER) \cite{DBLP:journals/corr/abs-1811-11682}, stores a fixed number of samples from the old tasks while learning the new task. Another, incremental Classifier and Representation Learning (iCaRL) \cite{icarl}, combines the LwF algorithm with some exemplars stored from the previously seen tasks. One generative rehearsal-based method is Deep Generative Replay \cite{DBLP:journals/corr/ShinLKK17}, where a copy of the generative model and classifier is stored and used to generate and label data from previous tasks, and the new labeled data is then added to the training data. This generative process is modified in Brain-Inspired Replay \cite{vandeVen2020} to handle more complex tasks by taking inspiration from the complementary learning systems of the mammalian brain. Other methods look at human-like memory-efficient techniques for storing and replaying samples and have been shown to mitigate catastrophic interference \cite{hafez2023map, ho2023prototype}.

A limitation of most rehearsal-based methods is that they require the calculation of complex measures to find exemplars to anchor certain weights to previous tasks and perform guided distribution shifts. These measures are often dependent on training parameter choices and loss calculations, making them harder to scale to new problem spaces. By contrast, the proposed power-law training environment-based rehearsal approach is model- and task-agnostic, making it easy to set up as a simulation for any task. 

\subsection*{CI mitigation techniques inspired from cognitive science literature}

Our rehearsal environment to observe the natural mitigation of CI is inspired by cognitive science research on how frequently humans experience previously seen tasks \cite{doi:10.1111/j.1467-9280.1991.tb00174.x, Heathcote2000} and the initial computational explorations by \citeauthor{lyndgaard2022catastrophic} (\citeyear{lyndgaard2022catastrophic}). ML researchers have previously explored mitigation techniques inspired by human behavior. \citeauthor{DavidsonMozer2020} (\citeyear{DavidsonMozer2020}) investigated how standard convolutional neural networks perform classification tasks when new visual tasks are introduced sequentially, simulating the way humans learn as they become experts in a particular domain. Another method, REplay using Memory INDexing (REMIND) \cite{remind}, looks at how the brain indexes memories, and explores efficient replay strategies with vector representations. Fearnet \cite{fearnet}, also inspired by the complementary learning systems of the mammalian brain, uses different networks for long-term and short-term memory to perform pseudo-rehearsal.

Surprisingly, no prior studies have explored the impact of naturalistic environmental structure on the learning of standard ML tasks. The current study simulates this structure to build a novel rehearsal methodology and shows the natural mitigation of CI.

\section{Methodology}
In this section, we describe the steps involved in simulating rehearsal methodologies inspired by the naturalistic learning environment of humans. Our work follows others \cite{van2022three} in comparing our approach to prevalent techniques on the SplitMNIST, SplitCIFAR-100, and SplitTinyImageNet task setups. The datasets for the experiment and the details of the training environment are described below.

\subsection{Class incremental learning scenario}
In this learning scenario, the model tries to incrementally learn a growing number of classes \cite{van2022three}. This differs from other types of incremental learning scenarios as the model has to both identify the task and also learn to discriminate between samples that are not observed together. Examples of Class Incremental Learning scenarios (Class-IL) are learning multiple object discrimination tasks or classification tasks (e.g., distinguishing between cats and dogs vs. distinguishing between cows and elephants) without seeing samples of all the different tasks at the same time (for example cats and cows). Previous work has shown that class incremental learning is the most difficult problem for continual learning in ML models \cite{van2022three, wang2023comprehensive}. 


\subsection{Datasets}
\label{datasets}

We show the mitigation properties of naturalistic environments by experimenting with three datasets that are prevalent in the continual learning and lifelong learning literature.

\subsubsection*{SplitMNIST protocol}
The MNIST dataset \cite{6296535} contains 60,000 training images and 10,000 test images, each a $28 \times 28$ pixel grayscale image of a single handwritten digit. For building the split MNIST protocol, this dataset was split into 5 contexts with each context having 2 randomly chosen digits. In each phase, we introduce a new context; see Figure \ref{fig:paper_class_cl_example}. The model learns to differentiate between digits while incrementally seeing new digits in different contexts.

\subsubsection*{SplitCIFAR-100 protocol}
The SplitCIFAR-100 dataset \cite{krizhevsky2009learning} contains 60,000 images of frequently seen objects in day-to-day life. There are 100 classes with each class having 500 training images and 100 test images, each a $32 \times 32$ colored image. The task follows the class-incremental task described by \cite{vandeven2019scenarios, DBLP:journals/corr/KirkpatrickPRVD16}, wherein each phase, we introduce a fixed number of new classes. Thus, the model learns to classify more classes with each new phase. Our experimental design ensures every phase has the same number of introduced classes and all classes are introduced by the end of the last phase.
The training data in a phase will have the maximum number of samples (500) for each of the newly introduced classes. For rehearsal-based approaches, the number of instances for each of the previous classes included in the training data depends upon the rehearsal type.

\subsubsection*{SplitTinyImageNet protocol}

The SplitTinyImageNet dataset \cite{Le2015TinyIV} consists of a subset of 100,000 images from the ImageNet visual classification challenge. There are 200 classes with each class having 500 training images and 50 test images, each $64 \times 64 \times 3$ pixels. The continual learning task formulation is similar to that of the SplitCIFAR-100 protocol. 

\begin{table*}[t]
    
\caption{Test accuracy after the final phase of training. The bracketed values indicate one $SD$ when running experiments with $N = 5$ seeds. LB: Lower Baseline, UB: Upper Baseline, EWC: Elastic Weight Consolidation, SI: Synaptic Intelligence, LwF: Learning without Forgetting, ER: Experience Replay, BIR: Brain-Inspired Replay, A-GEM: Averaged Gradient Episodic Memory, \textbf{PL: Power-law}, \textbf{Exp: Exponential}, iCaRL: Incremental Classifier and Representation Learning.}
\label{cl-comparions}
\vskip 0.15in
\begin{center}
\begin{small}
\begin{tabular}{llccccc}
\toprule
Type & Method & \multicolumn{2}{c}{SplitCIFAR-100} & \multicolumn{2}{c}{SplitTinyImageNet} & SplitMNIST\\ 
\multicolumn{2}{c}{Total number of phases:} & 10 Phases & 20 Phases & 10 Phases & 20 Phases & 5 Phases \\
\midrule
Baselines   & 
\multicolumn{6}{l}{
\begin{tabular}{@{}p{1.04cm}rrrrr@{}}
LB & 0.076 \small{(\textpm 0.005)} & 0.041 \small{(\textpm 0.003)} & 0.062 \small{(\textpm 0.002)} & 0.032 \small{(\textpm 0.001)} & 0.198 \small{(\textpm 0.001)}\\ UB &  \multicolumn{2}{c}{0.505 \small{(\textpm 0.002)}} &  \multicolumn{2}{c}{0.341 \small{(\textpm 0.004)}} & 0.978 \small{(\textpm 0.0)}
\end{tabular} 
}\\

\midrule
Parameter regularization   & 
\begin{tabular}{@{}l@{}}
EWC \\ SI
\end{tabular} &
\begin{tabular}{@{}c@{}}
0.079 \small{(\textpm 0.007)} \\ 0.081 \small{(\textpm 0.003)}

\end{tabular} 
& 
\begin{tabular}{@{}c@{}}
0.042 \small{(\textpm 0.002)} \\ 0.044 \small{(\textpm 0.002)}

\end{tabular} 
& 
\begin{tabular}{@{}c@{}}
0.059 \small{(\textpm 0.002)} \\ 0.058 \small{(\textpm 0.001)}

\end{tabular} 
& 
\begin{tabular}{@{}c@{}}
0.034 \small{(\textpm 0.001)} \\ 0.033 \small{(\textpm 0.001)}

\end{tabular} 
&
\begin{tabular}{@{}c@{}}
0.231 \small{(\textpm 0.046)} \\ 0.208 \small{(\textpm 0.016)}

\end{tabular}
\\
\midrule
Functional regularization 
& LwF
& 0.255 \small{(\textpm 0.009)}
& 0.087 \small{(\textpm 0.005)}
& 0.131 \small{(\textpm 0.004)}
& 0.067 \small{(\textpm 0.001)}
& 0.212 \small{(\textpm 0.005)}
\\
\midrule
Rehearsal methods   & 
\begin{tabular}{@{}l@{}}
ER \\ PL \\Exp \\ BIR \\ A-GEM 
\end{tabular} &
\begin{tabular}{@{}c@{}}
0.399 \small{(\textpm 0.006)} 
\\ \textbf{\textit{0.429 \small{(\textpm 0.006)}}}  
\\ 0.373 \small{(\textpm 0.006)}
\\ 0.222 \small{(\textpm 0.011)} 
\\ 0.276 \small{(\textpm 0.012)} 

\end{tabular} &
\begin{tabular}{@{}c@{}}
0.386 \small{(\textpm 0.004)} 
\\  \textbf{\textit{0.432 \small{(\textpm 0.007)}}}
\\ 0.359 \small{(\textpm 0.008)}
\\ 0.253 \small{(\textpm 0.012)} 
\\ 0.214 \small{(\textpm 0.046)} 

\end{tabular} &
\begin{tabular}{@{}c@{}}
0.242 \small{(\textpm 0.002)} 
\\ \textbf{\textit{0.25 \small{(\textpm 0.009)}}}  
\\ 0.222 \small{(\textpm 0.007)}
\\ 0.083 \small{(\textpm 0.007)} 
\\ 0.062 \small{(\textpm 0.003)} 

\end{tabular} &
\begin{tabular}{@{}c@{}}
0.207 \small{(\textpm 0.006)}  
\\  \textbf{\textit{0.27 \small{(\textpm 0.003)}}}
\\ 0.183 \small{(\textpm 0.003)}
\\ 0.065 \small{(\textpm 0.003)} 
\\ 0.036 \small{(\textpm 0.002)} 

\end{tabular} &
\begin{tabular}{@{}c@{}}
\textbf{0.965 \small{(\textpm 0.002)}} 
\\  \textit{0.959 \small{(\textpm 0.003)}}
\\ 0.948 \small{(\textpm 0.002)}
\\ 0.944 \small{(\textpm 0.006)} 
\\ 0.7 \small{(\textpm 0.219)} 

\end{tabular}
\\
\midrule
Template based   & 
 iCaRL
 & 0.329 \small{(\textpm 0.01)}
 & 0.186 \small{(\textpm 0.007)}
 & 0.068 \small{(\textpm 0.014)}
 & 0.005 \small{(\textpm 0.002)}
 & 0.943 \small{(\textpm 0.003)}
\\

\bottomrule
\end{tabular}
\end{small}
\end{center}
\end{table*}
\subsection {Model architecture} 

Following prior work \cite{van2022three}, we use model architectures that are specific to each protocol formulation. For a fair comparison between the learning environments and the alternative methods presented in Table \ref{cl-comparions}, the same model architecture is used for experiments on each dataset. Finally, in all of our experiments, we replicate results with 5 seeds to ensure reproducibility and robustness to the initialization of weights in a network. The details of the model architecture are included in the supplementary materials.

\subsection{Naturalistic environments}
Inspired by naturalistic environments \cite{doi:10.1111/j.1467-9280.1991.tb00174.x, Heathcote2000}, we introduce two function-based rehearsal environments: power-law and exponential. These functions determine the number of samples to rehearse of previously seen classes. For all rehearsals, we perform \textbf{simple random sampling}: the number of samples of a particular task to be rehearsed in each phase is randomly chosen from the samples of the task that were in the previous phase. That is, the rehearsal set is not resampled from scratch at each new phase. Rather, some samples of previous tasks from the prior phase are randomly dropped at the new phase.

\subsubsection{Power-law distribution:}
Each training phase of this setup has the maximum number of samples for the newly introduced classes, while the data distribution for the older classes introduced in previous phases follows a decreasing power function such that the number of instances of the classes introduced in the first phase will have the least representation in the last phase. This follows naturalistic human learning where humans see rehearsals in the environment at a decreasing power function frequency. The power-law sample distribution equation for a class is given by equation (\ref{eq:pwl}), where the value for $x$ is the number of phases passed after the introduction of the class.
\begin{equation}
f(x) = ax^{-b}
\label{eq:pwl}
\end{equation}

\subsubsection{Exponential distribution:}
Similar to the power-law, each training phase of this setup has the maximum number of samples for the newly introduced classes, while the data distribution for the older classes introduced in previous phases follows a decreasing exponential function. 
The exponential sample distribution equation for a class is given by equation (\ref{eq:exp}), where the value for $x$ is the same as for equation (\ref{eq:pwl}).

\begin{equation}
f(x) = ae^{-xb}
\label{eq:exp}
\end{equation}

\subsection{Training and evaluation}
The environment (i.e., power-law, exponential, etc.) determines the distribution of samples for rehearsal during the training phases. We challenge existing mitigation methods by showing that the rehearsal of old tasks in natural environments comparably mitigates forgetting. This implies that complex budgeting or regularization algorithms may not be needed for continual learning in many situations.

The models are tested on uniformly distributed held-out data. This is similar to conventional training and testing splits, however, the test set only consists of classes that have been included in the training set for at least one phase. The standard uniform test distribution was chosen to equally weight performance by classifying all classes equally and to enable comparisons of the intrinsic environment-based mitigation to existing techniques for mitigating CI.

\noindent \textbf{Comparison to existing CL methods:} We compare the CI mitigation properties of the naturalistic learning environments to the representative methods from prior literature and the baselines described below (Table \ref{cl-comparions}). In all rehearsal-based methods (ER, PL, Exp, AGEM), our operationalization ensures that each model cumulatively rehearses the same number of samples by the last phase, i.e., our experiments equate the number of total rehearsals across all the rehearsal-based methods. We do not modify the implementation details of other approaches from \citet{van2022three}. The individual details of the methods are described in the supplementary materials.

Beyond methods from the prior literature, the work uses the following as baselines to show the efficacy of a power-law training environment.

\begin{itemize}    
    \item \textbf{Lower-baseline}: We establish a \emph{lower-bound performance baseline} for our datasets. In this training setup, each training phase only has access to the data for the newly introduced task. In particular, there is \emph{no rehearsal of previously seen classes}. This is akin to a regular neural network architecture which sequentially learns new tasks. This environment is expected to show high levels of CI.
    \item \textbf{Upper-baseline}: In this baseline case, we establish an \emph{upper-bound performance baseline} for our datasets. In this setup, the model is trained over data from all classes in a single phase. This is a non-continual learning scenario represented by the flat line in Figure \ref{fig:paper_fig_1}.
\end{itemize}
\subsection{Combining naturalistic rehearsal environments with regularization methods}
\label{sec:combination}
\begin{table*}[t]
    
\caption{Test accuracies after the final phase of training in combination approaches. The bracketed values indicate one $SD$ when running experiments with $N = 5$ seeds. EWC: Elastic Weight Consolidation, SI: Synaptic Intelligence, LwF: Learning without Forgetting, ER: Experience Replay, \textbf{PL: Power-law}, \textbf{Exp: Exponential}.}
\label{tab:combination}
\begin{center}
\begin{small}
\begin{tabular}{llccc}
\toprule
Rehearsal & Regularization & SplitCIFAR-100 & SplitTinyImageNet & SplitMNIST\\ 
\midrule



ER   & 
\begin{tabular}{@{}l@{}}
None \\ EWC \\ SI \\ LwF  
\end{tabular} &
\begin{tabular}{@{}c@{}}
0.399 \small{(\textpm 0.006)} 
\\ 0.424 \small{(\textpm 0.004)} 
\\ 0.404 \small{(\textpm 0.004)} 
\\  \textbf{0.457 \small{(\textpm 0.004)}}  
\end{tabular}
&
\begin{tabular}{@{}c@{}}
0.242 \small{(\textpm 0.002)} 
\\ 0.203 \small{(\textpm 0.003)} 
\\ 0.201 \small{(\textpm 0.004)} 
\\  \textbf{0.235 \small{(\textpm 0.004)}}  
\end{tabular}
&
\begin{tabular}{@{}c@{}}
0.965 \small{(\textpm 0.002)} 
\\ 0.926 \small{(\textpm 0.011)} 
\\ 0.832 \small{(\textpm 0.087)} 
\\  \textbf{0.965 \small{(\textpm 0.004)}}  
\end{tabular}
\\
\midrule
PL   & 
\begin{tabular}{@{}l@{}}
None \\ EWC \\ SI \\ LwF  
\end{tabular} &
\begin{tabular}{@{}c@{}}
0.429 \small{(\textpm 0.006)}
\\ 0.424 \small{(\textpm 0.005)} 
\\ 0.417 \small{(\textpm 0.004)} 
\\  \textbf{0.461 \small{(\textpm 0.007)}}  
\end{tabular}
 &
\begin{tabular}{@{}c@{}}
0.25 \small{(\textpm 0.009)}
\\ 0.25 \small{(\textpm 0.003)} 
\\ 0.27 \small{(\textpm 0.004)} 
\\  \textbf{0.289 \small{(\textpm 0.002)}}  
\end{tabular}
&
\begin{tabular}{@{}c@{}}
0.959 \small{(\textpm 0.003)} 
\\ 0.882 \small{(\textpm 0.074)} 
\\ 0.803 \small{(\textpm 0.06)} 
\\  \textbf{0.964 \small{(\textpm 0.002)}}  
\end{tabular}
\\
\midrule
Exp   & 
\begin{tabular}{@{}l@{}}
None \\ EWC \\ SI \\ LwF  
\end{tabular} &
\begin{tabular}{@{}c@{}}
0.373 \small{(\textpm 0.006)}
\\ 0.387 \small{(\textpm 0.012)} 
\\ 0.374 \small{(\textpm 0.011)} 
\\  \textbf{0.449 \small{(\textpm 0.006)}}  
\end{tabular}
&
\begin{tabular}{@{}c@{}}
0.222 \small{(\textpm 0.007)}
\\ 0.251 \small{(\textpm 0.008)} 
\\ 0.247 \small{(\textpm 0.004)} 
\\  \textbf{0.272 \small{(\textpm 0.003)}}  
\end{tabular}
&
\begin{tabular}{@{}c@{}}
0.948 \small{(\textpm 0.002)} 
\\ 0.875 \small{(\textpm 0.072)} 
\\ 0.82 \small{(\textpm 0.048)} 
\\  \textbf{0.96 \small{(\textpm 0.003)}}  
\end{tabular}
\\
\bottomrule
\end{tabular}
\end{small}
\end{center}
\end{table*}
The proposed rehearsal environments are orthogonal to several other prevalent methods discussed in the literature. This enables us to combine the different rehearsal environments (Experience Replay, Power-law, Exponential) with the different regularization methods (LwF, EWC, and SI). In each phase, the rehearsal environment determines the number of samples seen by the model while the regularization methods operate on the loss functions and optimizers. (Note: We cannot combine template-based methods like iCaRL with rehearsal environments because those methods already store and replay data for template creations. Similarly, environment-based rehearsal methods cannot be combined with generative rehearsal methods as the generative model serves as a substitute for a replay buffer.)


\section{Results}

\subsection*{What is the extent of CI mitigation with natural rehearsal in simulated power-law training environments?}
For the SplitMNIST, SplitCIFAR-100, and SplitTinyImageNet datasets, we investigate the natural mitigation of CI in the power-law and exponential rehearsal environments by comparing the test accuracy of all the classes at the end of each phase. We compare this performance to that of the baseline approaches and popular mitigation methods.

For the SplitMNIST data, Table \ref{cl-comparions} shows the average test accuracy over all classes after 5 phases. For SplitCIFAR-100 and SplitTinyImageNet, we also test whether increasing the number of phases from 10 to 20 (i.e., while decreasing the number of classes introduced per phase) leads to a drop in mitigation performance for all training environments. Additionally, Figure \ref{fig:paper_fig_1} shows the average test accuracy for all introduced classes until a phase number ($n$) for SplitCIFAR-100 and SplitTinyImageNet.

We make the following observations about the findings in Table \ref{cl-comparions}:

\begin{itemize}
    \setlength\itemsep{0em}
    \setlength\parskip{1em}
    \setlength\parsep{0em}
\item \textbf{Environment-based rehearsal methods (ER, Power-law, and Exponential) that have access to previous data in different distributions (i.e., for old tasks) perform better than methods where no such data are included.}
\item For the simpler SplitMNIST task, almost all rehearsal and template-based methods show near-ceiling performance. Here, experience replay performs slightly better than power-law.
\item For the much harder SplitCIFAR-100 and SplitTinyImageNet tasks, \emph{power-law performs descriptively better compared to other training environments} with the expected exception of the upper baseline.

\item For the SplitCIFAR-100 and SplitTinyImageNet tasks, increasing the number of phases (i.e., while simultaneously decreasing the number of classes introduced per phase) leads to a greater advantage for power-law compared to other techniques.
\end{itemize}
Our results provide strong empirical evidence for natural mitigation of CI in rehearsal environments, \textbf{with performance that exceeds existing methods}. These results show human-like ability to remember previously seen classes while learning new ones when the distribution of the class samples follows a power-law.

\subsection*{When combined with different rehearsal environments, how well do prevalent regularization approaches mitigate CI?}

Regularization methods add a penalty term in the loss function to ensure that the gradient updates result in less catastrophic forgetting for the previous classes. We explore the performance of these regularization methods when used in conjunction with different rehearsal environments. Table \ref{tab:combination} shows that parameter regularization methods do \emph{not} generally improve performance above and beyond rehearsal environments in isolation. However, there is one important exception: \textbf{LwF shows improved CI mitigation in all rehearsal environments} (SplitMNIST: $\text{Acc}_{\text{PL, LwF}} -\text{Acc}_{\text{PL, None}} = 0.005$, SplitCIFAR-100: $\text{Acc}_{\text{PL, LwF}} - \text{Acc}_{\text{PL, None}} = 0.032$, SplitTinyImageNet: $\text{Acc}_{\text{PL, LwF}} - \text{Acc}_{\text{PL, None}} = 0.039$) 

We attribute this pattern of results to the training loss objectives of the three regularization methods. In EWC and SI, while training in the current phase, parameter regularization restricts weight changes for parameters that are important for previously seen classes. 
However, in rehearsal-based methods, the accumulation of data from previous tasks in the current training corpus results in the loss term updating fewer parameters for new classes. Thus, with more phases, the parameter regularization term becomes less effective. By contrast, in LwF, the network is encouraged to not change at certain anchor points. Thus, adding functional regularization does improve the performance. 

\begin{table*}[h]
\caption{Test accuracy on SplitCIFAR-100 as a function of the minimum proportion of samples for each previous task in the last phase of power-law rehearsal environments. Experience replay and exponential rehearsal environments are balanced to match the same number of total samples rehearsed as the power-law environment. The bracketed values indicate one $SD$ when running experiments with $N = 5$ seeds. ER: Experience Replay, \textbf{PL: Power-law}, \textbf{Exp: Exponential}}
\label{minimum_data_size_both_tasks}
\vskip 0.15in
\begin{center}
\begin{tabular}{lccccc}
\toprule
Dataset & Distribution & \multicolumn{4}{c}{Minimum proportion of samples}\\ 
& & 1\% & 2\% & 5\% & 10\% \\
\midrule
& ER & 0.264 \small{(\textpm 0.005)} & 0.292 \small{(\textpm 0.004)} & 0.324 \small{(\textpm 0.004)} & 0.399 \small{(\textpm 0.006)} \\
SplitCIFAR-100 & PL & 0.289 \small{(\textpm 0.01)} & 0.359 \small{(\textpm 0.036)} & 0.392 \small{(\textpm 0.007)} & 0.432 \small{(\textpm 0.007)} \\
& Exp & 0.223 \small{(\textpm 0.013)} & 0.237 \small{(\textpm 0.012)} & 0.308 \small{(\textpm 0.011)} & 0.373 \small{(\textpm 0.006)} \\
\midrule
& ER & 0.171 \small{(\textpm 0.006)} & 0.194 \small{(\textpm 0.003)} & 0.22 \small{(\textpm 0.002)} & 0.242 \small{(\textpm 0.002)} \\
SplitTinyImageNet & PL & 0.161 \small{(\textpm 0.002)} & 0.193 \small{(\textpm 0.004)} & 0.241 \small{(\textpm 0.004)} & 0.25 \small{(\textpm 0.009)} \\
& Exp & 0.141 \small{(\textpm 0.002)} & 0.152 \small{(\textpm 0.001)} & 0.192 \small{(\textpm 0.007)} & 0.222 \small{(\textpm 0.007)} \\

\bottomrule
\end{tabular}
\end{center}
\vskip -0.1in
\end{table*}

\subsection*{Generalization and robustness checks}

We conducted multiple experiments (ablation studies) to understand the generalizability and robustness of our rehearsal environments in mitigating CI. We provide the operational details of these experiments in the supplementary materials. Brief descriptions along with the key takeaways of the different experiments are as follows:
\begin{itemize}
    \setlength\itemsep{0em}
    \setlength\parskip{1em}
    \setlength\parsep{0em}

    \item \textbf{Minimum proportions of samples required per class:} In power-law and exponential learning environments, classes introduced in phase 1 have the least representation of the rehearsal data in all subsequent phases. This proportion also determines the constants in the power-law function curves. We vary the minimum amount of rehearsal data required for previous classes to observe whether the mitigation properties of the power-law rehearsal environment are maintained. We compare the power-law environment to the Experience Replay and the exponential rehearsal environment by matching the total number of rehearsed samples with the power-law curve. For the SplitCIFAR-100 and SplitTinyImageNet data, there is a steady decline in test accuracy when varying the minimum proportion of samples with different sample percentages. \textit{However, we find that power-law rehearsal environments outperform other environment-based rehearsal methods (ER, Exp) across almost all variations in the minimum number of samples of a class in any phase} (refer to Table \ref{minimum_data_size_both_tasks}).
    \item \textbf{Varying the number of hidden layers in the model:} Increasing the number of fully connected hidden layers can affect the performance of models trained in different environments. The correlation between model complexity (size) and accuracy is not always direct and positive. \emph{The performance of the power-law rehearsal environment is better than all other model architectures and all mitigation techniques for all model complexities} (more information in the supplementary materials). 
    \item \textbf{Varying the total size of training data for each class:} Variation in the total size of training data for each class increases the problem complexity. \emph{Naturalistic environments (Exp, PL) have the least drop in performance with an increase in problem complexity} (refer to the supplementary materials).
    \item \textbf{Domain-Incremental Learning Scenario:} In this learning scenario, the problem structure has a shift in the input distribution but has the same possible output space as the previous tasks \cite{van2022three}. Although these tasks are not as difficult as Class incremental tasks for ML models, they remain the most popular forms of continual learning problems in practice \cite{dai2023domain, van2022three}. \emph{Our experiments for power-law rehearsal environment simulation in Domain-IL tasks show strong mitigation behavior.}
\end{itemize}

\section{Conclusion}
We apply the naturalistic learning environments that humans experience to the continual learning problem in neural networks. Our simulations offer an evaluation of sequential learning capabilities and are realistic in accounting for the rehearsal of data in the real world.

We simulate model training in different environments on three Class-Incremental Learning scenarios:  MNIST handwritten digit recognition tasks and the CIFAR 100 and TinyImageNet object classification tasks. The networks are trained on multiple phases to understand the extent of CI in different setups. We show that the intrinsic mitigation in the rehearsal environments performs comparably to prevalent CI mitigation techniques in the literature, suggesting that in domains with a power-law distribution of tasks, complex CI mitigation strategies may not be required. Most importantly, the power-law rehearsal environment shows state-of-the-art accuracy on the SplitCIFAR-100 tasks. These models behave like humans in showing less forgetting of previous tasks as they continue to encounter new tasks in a decreasing power-law manner. Furthermore, we explore combining rehearsal environments with other types of approaches and show that functional regularization methods (i.e., LwF) in conjunction with rehearsal environments further increase performance. Finally, the environments are robust to variations in the minimum proportion of samples seen, the number of layers of the model, the total number of samples of each class, and the learning scenario types.

Overall, our experiments demonstrate that the naturalistic power-law training environment mitigates CI better than most other methods in the literature. As such, it provides a new standard for future research in continual learning.

\section{Experimental Infrastructure}
All experiments were conducted with PyTorch \cite{pytorch} on NVIDIA RTX 6000 GPUs with 24GB GPU RAM. 
\section{Limitations and Future Work}
Naturally occurring frequency distributions are proposed to empirically follow power-law curves, which, in reality, are "power-law \emph{like}" stochastic distributions that contain random spikes and lows. We simulate each class to have a perfect power-law distribution which may not accurately represent naturally occurring data. Furthermore, while empirical evidence shows that many naturally occurring situations follow a power-law frequency distribution, they do not represent all possible naturally occurring situations. Future work can explore simulating stochastic naturalistic environments.

While we explore mitigation of CI for three popular datasets, future work can explore more diverse problems containing multiple visual properties of an image, for example, color and texture-based images as found in the CLEVR dataset \cite{johnson2016clevr}. The models utilized to solve these problems would also be more complex Vision Transformer-based \cite{dosovitskiy2021image} architectures and might provide more insight into the capabilities and drawbacks of the natural mitigation of a power-law training environment. 

The paper compares naturalistic rehearsal environments to representative techniques from prior literature. While these techniques are highly prevalent and used for comparison, they do not comprehensively cover all techniques for the mitigation of CI. Future work can compare our environments with additional techniques, and also combine rehearsal environments with them.

The results show that the benefit of naturalistic environments becomes more pronounced as the number of phases increases. However, we did not investigate whether this advantage continues to grow with increasing task complexity, which can be explored in future research.

For all the experiments reported in the paper, we use the model architecture and the respective hyper-parameters based on previous work \cite{van2022three}. This means we fix the dropout probability, learning rate, batch size, maximum training iterations per task, and number of units per layer. We do not carry out experiments with different loss functions and optimizers. Also, we do not implement early stopping. The results can be somewhat different if we use optimal parameters obtained from grid search; however, our current results show that the difference between the performance of power-law and other approaches is significant even when the sample size and the model complexity are varied.

\bibliography{citations}
\newpage
\onecolumn

\section{Supplementary Materials}

\subsection{Domain Incremental Learning scenario}
In this learning scenario, the problem structure has a shift in the input distribution but has the same possible output space as the previous tasks \cite{van2022three}. Examples of Domain Incremental Learning scenarios (Domain-IL) are object detection tasks or classification tasks with a change in contexts; like detecting polar bears in images with the background of mountains vs. floating ice bodies or finding sarcasm on different platforms such as Reddit and Twitter. While Domain-IL tasks are not as difficult as Class incremental tasks for ML models, they remain the most popular forms of continual learning problems in practice \cite{van2022three, dai2023domain}. We ablate on the Domain-IL scenario because of this real-world demand. We use a representative task formulation for the Domain incremental learning scenario, the permuted MNIST task. For building the permuted MNIST task, we add padding to the images in the MNIST dataset to rescale them to 32x32 pixel images and then generate a fixed, random permutation by which the input pixels of all images are shuffled. The model learns to identify the number corresponding to the digit despite the permutation. Each task is associated with a characteristic permutation, and the same permutation is applied across the entire dataset. The tasks formed by the transformations follow the domain-incremental learning task described by \cite{vandeven2019scenarios, DBLP:journals/corr/KirkpatrickPRVD16}, where the task identity is not available during performance.

\begin{table*}[h]
\centering
 \caption{  Test accuracy in the final phase as we increase the number of phases in the \emph{Domain Incremental Learning scenario}. The bracketed values indicate one $SD$ when running experiments with $N = 5$ seeds.}
  \resizebox{0.43\textwidth}{!}{
  \begin{tabular}{lcc}
    \hline
    Training & \multicolumn{2}{c}{Avg test accuracy}\\
    environment & 10 phases & 20 phases\\
    
    \hline
    
    Upper baseline & 0.936 \small{(\textpm 0.003)} & 0.874 \small{(\textpm 0.003)} \\
     Power-law & 0.901 \small{(\textpm 0.004)} & 0.851 \small{(\textpm 0.004)} \\
     EWC & 0.743 \small{(\textpm 0.024)} & 0.523 \small{(\textpm 0.016)} \\
     Lower baseline & 0.472 \small{(\textpm 0.007)} & 0.424 \small{(\textpm 0.012)}\\
    \hline

\end{tabular}
}
 
  \label{tab:domain_IL}
\end{table*}

\subsection{Minimum proportions of samples required per class}

In the SplitCIFAR-100 and SplitTinyImagenet data, for every phase after Phase 2, classes introduced in Phase 1 have the smallest representation in the rehearsal data. By varying this, we test the minimum amount of rehearsal data required for a model to maintain the mitigation of CI. Most experiments presented in this paper for the power-law environment assume that the minimum number of samples in the last phase for classes introduced in phase 1 is 10\% of the original training size of each task. In this section, we explore how varying this minimum data value affects the performance of models trained in a power-law setup. 

Table \ref{minimum_data_size} shows the performance of the power-law models after training over 10 phases when varying the minimum number of samples with four different sample percentages. Only when the minimum number drops below 2\% of the total class sample size is there a considerable decline in the performance of the model. At 1\%, model accuracy begins to collapse.

\begin{table*}[h]
\caption{SplitCIFAR-100: Test accuracy as a function of the minimum proportion of samples for each previous task in the last phase of power-law rehearsal environments. Experience Replay and exponential rehearsal environments are balanced to match the same number of total samples rehearsed as the power-law environment. The bracketed values indicate one $SD$ when running experiments with $N = 5$ seeds. ER: Experience Replay, \textit{PL: Power-law}, \textit{Exp: Exponential}}
\label{minimum_data_size}
\vskip 0.15in
\begin{center}
\begin{tabular}{lcccc}
\toprule
Distribution & \multicolumn{4}{c}{Minimum proportion of samples}\\ 
& 1\% & 2\% & 5\% & 10\% \\
\midrule
ER & 0.264 \small{(\textpm 0.005)} & 0.292 \small{(\textpm 0.004)} & 0.324 \small{(\textpm 0.004)} & 0.399 \small{(\textpm 0.006)} \\
PL & 0.289 \small{(\textpm 0.01)} & 0.359 \small{(\textpm 0.036)} & 0.392 \small{(\textpm 0.007)} & 0.432 \small{(\textpm 0.007)} \\
Exp & 0.223 \small{(\textpm 0.013)} & 0.237 \small{(\textpm 0.012)} & 0.308 \small{(\textpm 0.011)} & 0.373 \small{(\textpm 0.006)} \\

\bottomrule
\end{tabular}
\end{center}
\vskip -0.1in
\end{table*}

\begin{table*}[h]
\caption{SplitTinyImagenet: Test accuracy as a function of the minimum proportion of samples for each previous task in the last phase of power-law rehearsal environments. Experience Replay and exponential rehearsal environments are balanced to match the same number of total samples rehearsed as the power-law environment. The bracketed values indicate one $SD$ when running experiments with $N = 5$ seeds. ER: Experience Replay, \textit{PL: Power-law}, \textit{Exp: Exponential}}
\label{minimum_data_size_tiny_image_net}
\vskip 0.15in
\begin{center}
\begin{tabular}{lcccc}
\toprule
Distribution & \multicolumn{4}{c}{Minimum proportion of samples}\\ 
& 1\% & 2\% & 5\% & 10\% \\
\midrule
ER & 0.171 \small{(\textpm 0.006)} & 0.193 \small{(\textpm 0.003)} & 0.22 \small{(\textpm 0.002)} & 0.242 \small{(\textpm 0.002)} \\
PL & 0.161 \small{(\textpm 0.002)} & 0.193 \small{(\textpm 0.004)} & 0.241 \small{(\textpm 0.004)} & 0.25 \small{(\textpm 0.009)} \\
Exp & 0.141 \small{(\textpm 0.002)} & 0.152 \small{(\textpm 0.001)} & 0.192 \small{(\textpm 0.007)} & 0.222 \small{(\textpm 0.007)} \\

\bottomrule
\end{tabular}
\end{center}
\vskip -0.1in
\end{table*}

\subsection{Varying the number of hidden layers in the model (SplitCIFAR-100 and SplitTinyImageNet protocols)}

Increasing the number of fully connected hidden layers can affect the performance of models trained in different environments. The correlation between model complexity (size) and accuracy is not always direct and positive; it also depends upon the difficulty of the tasks. Here, we investigate whether model complexity affects performance and whether different model complexities continue to display CI mitigation. Tables \ref{cl-comparions-hidden-layer} and \ref{cl-comparions-hidden-layer-tiny} show the performance of the models with different numbers of hidden layers. Note that the performance of the rehearsal methods in the power-law environment is better than all other model architectures and all mitigation techniques for all model complexities. 

\begin{table*}[h]
    
\caption{SplitCIFAR-100: Test accuracy in the final phase as we change the number of fully connected hidden layers in the model. The bracketed values indicate one $SD$ when running experiments with $N = 5$ seeds. LB: Lower Baseline, UB: Upper Baseline, EWC: Elastic Weight Consolidation, SI: Synaptic Intelligence, LwF: Learning without Forgetting, ER: Experience Replay, BIR: Brain-Inspired Replay, A-GEM: Averaged Gradient Episodic Memory, \textit{PL: Power-law}, \textit{Exp: Exponential}, iCaRL: Incremental Classifier and Representation Learning, DGR: Deep Generative Replay}
\label{cl-comparions-hidden-layer}
\vskip 0.15in
\begin{center}
\begin{small}
\begin{tabular}{llccccc}
\toprule
Type & Method & \multicolumn{5}{c}{No. of hidden layers}\\ 
& & 1 & 2 & 3 & 4 & 5\\
\midrule
Baselines   & 
\begin{tabular}{@{}l@{}}
LB \\ UB
\end{tabular} &
\begin{tabular}{@{}c@{}}
0.073 \small{(\textpm 0.004)}
\\ 
0.39 \small{(\textpm 0.003)}
\end{tabular}
&
\begin{tabular}{@{}c@{}}
0.075\small{(\textpm 0.005)}
\\ 
0.45 \small{(\textpm 0.003)}
\end{tabular}
&
\begin{tabular}{@{}c@{}}
0.076 \small{(\textpm 0.005)}
\\ 
0.505 \small{(\textpm 0.002)}
\end{tabular}
&
\begin{tabular}{@{}c@{}}
0.071 \small{(\textpm 0.005)}
\\ 
0.444 \small{(\textpm 0.003)}
\end{tabular}
&
\begin{tabular}{@{}c@{}}
0.068 \small{(\textpm 0.005)}
\\ 
0.411 \small{(\textpm 0.003)}
\end{tabular}
\\
\midrule
Parameter Regularization   & 
\begin{tabular}{@{}l@{}}
EWC \\ SI
\end{tabular} &
\begin{tabular}{@{}c@{}}
0.116 \small{(\textpm 0.008)} \\ 0.111 \small{(\textpm 0.007)}

\end{tabular} 
& 
\begin{tabular}{@{}c@{}}
0.077 \small{(\textpm 0.003)}\\ 0.081 \small{(\textpm 0.002)}

\end{tabular}
& 
\begin{tabular}{@{}c@{}}
0.079 \small{(\textpm 0.007)} \\ 0.081 \small{(\textpm 0.003)}

\end{tabular}
& 
\begin{tabular}{@{}c@{}}
0.07 \small{(\textpm 0.005)} \\ 0.069 \small{(\textpm 0.005)}
\end{tabular}
& 
\begin{tabular}{@{}c@{}}
0.066 \small{(\textpm 0.003)} \\ 0.066 \small{(\textpm 0.005)}
\end{tabular}
\\
\midrule
Functional Regularization 
& LwF
& 0.089 \small{(\textpm 0.003)}
& 0.2 \small{(\textpm 0.01)}
& 0.255 \small{(\textpm 0.009)}
& 0.189 \small{(\textpm 0.005)}
& 0.137 \small{(\textpm 0.007)}
\\
\midrule
Rehearsal methods   & 
\begin{tabular}{@{}l@{}}
ER \\ PL \\Exp \\ BIR \\ A-GEM
\end{tabular} &
\begin{tabular}{@{}c@{}}
0.367 \small{(\textpm 0.003)} 
\\ \textbf{0.372 \small{(\textpm 0.005)}}  
\\ 0.31 \small{(\textpm 0.007)}
\\ 0.081 \small{(\textpm 0.003)} 
\\ 0.127 \small{(\textpm 0.008)} 

\end{tabular} &
\begin{tabular}{@{}c@{}}
0.36 \small{(\textpm 0.004)} 
\\  \textbf{0.418 \small{(\textpm 0.008)}} 
\\ 0.361 \small{(\textpm 0.01)}
\\ 0.231 \small{(\textpm 0.007)} 
\\ 0.178 \small{(\textpm 0.007)} 

\end{tabular} &
\begin{tabular}{@{}c@{}}
0.399 \small{(\textpm 0.006)} 
\\  \textbf{0.429 \small{(\textpm 0.006)}} 
\\ 0.373 \small{(\textpm 0.006)}
\\ 0.222 \small{(\textpm 0.011)} 
\\ 0.276 \small{(\textpm 0.012)} 

\end{tabular} &
\begin{tabular}{@{}c@{}}
0.283 \small{(\textpm 0.004)} 
\\  \textbf{0.355 \small{(\textpm 0.018)}} 
\\ 0.305 \small{(\textpm 0.015)}
\\ 0.171 \small{(\textpm 0.013)} 
\\ 0.117 \small{(\textpm 0.021)} 

\end{tabular} &
\begin{tabular}{@{}c@{}}
0.264 \small{(\textpm 0.007)} 
\\  \textbf{0.335 \small{(\textpm 0.01)}} 
\\ 0.282 \small{(\textpm 0.014)}
\\ 0.139 \small{(\textpm 0.01)} 
\\ 0.08 \small{(\textpm 0.007)} 

\end{tabular}
\\
\midrule
Template Based   & 
 iCaRL
 & 0.251 \small{(\textpm 0.003)}
 & 0.212 \small{(\textpm 0.003)}
 & 0.329 \small{(\textpm 0.01)}
 & 0.302 \small{(\textpm 0.005)}
 & 0.27 \small{(\textpm 0.007)}
\\

\bottomrule
\end{tabular}
\end{small}
\end{center}
\end{table*}

\begin{table*}[h]
    
\caption{SplitTinyImagenet: Test accuracy in the final phase as we change the number of fully connected hidden layers in the model. The bracketed values indicate one $SD$ when running experiments with $N = 5$ seeds. LB: Lower Baseline, UB: Upper Baseline, EWC: Elastic Weight Consolidation, SI: Synaptic Intelligence, LwF: Learning without Forgetting, ER: Experience Replay, BIR: Brain-Inspired Replay, A-GEM: Averaged Gradient Episodic Memory, \textit{PL: Power-law}, \textit{Exp: Exponential}, iCaRL: Incremental Classifier and Representation Learning, DGR: Deep Generative Replay}
\label{cl-comparions-hidden-layer-tiny}
\vskip 0.15in
\begin{center}
\begin{small}
\begin{tabular}{llccccc}
\toprule
Type & Method & \multicolumn{5}{c}{No. of hidden layers}\\ 
& & 1 & 2 & 3 & 4 & 5\\
\midrule
Baselines   & 
\begin{tabular}{@{}l@{}}
LB \\ UB
\end{tabular} &
\begin{tabular}{@{}c@{}}
0.057 \small{(\textpm 0.001)}
\\ 
0.282 \small{(\textpm 0.005)}
\end{tabular}
&
\begin{tabular}{@{}c@{}}
0.062\small{(\textpm 0.001)}
\\ 
0.3 \small{(\textpm 0.004)}
\end{tabular}
&
\begin{tabular}{@{}c@{}}
0.062 \small{(\textpm 0.002)}
\\ 
0.341 \small{(\textpm 0.004)}
\end{tabular}
&
\begin{tabular}{@{}c@{}}
0.062 \small{(\textpm 0.002)}
\\ 
0.284 \small{(\textpm 0.011)}
\end{tabular}
&
\begin{tabular}{@{}c@{}}
0.062 \small{(\textpm 0.002)}
\\ 
0.26 \small{(\textpm 0.011)}
\end{tabular}
\\
\midrule
Parameter Regularization   & 
\begin{tabular}{@{}l@{}}
EWC \\ SI
\end{tabular} &
\begin{tabular}{@{}c@{}}
0.058 \small{(\textpm 0.007)} \\ 0.059 \small{(\textpm 0.004)}

\end{tabular} 
& 
\begin{tabular}{@{}c@{}}
0.061 \small{(\textpm 0.002)}\\ 0.061 \small{(\textpm 0.003)}

\end{tabular}
& 
\begin{tabular}{@{}c@{}}
0.059 \small{(\textpm 0.002)} \\ 0.058 \small{(\textpm 0.001)}

\end{tabular}
& 
\begin{tabular}{@{}c@{}}
0.058 \small{(\textpm 0.005)} \\ 0.061 \small{(\textpm 0.004)}
\end{tabular}
& 
\begin{tabular}{@{}c@{}}
0.06 \small{(\textpm 0.004)} \\ 0.057 \small{(\textpm 0.002)}
\end{tabular}
\\
\midrule
Functional Regularization 
& LwF
& 0.085 \small{(\textpm 0.001)}
& 0.145 \small{(\textpm 0.005)}
& 0.131 \small{(\textpm 0.004)}
& 0.064 \small{(\textpm 0.0)}
& 0.055 \small{(\textpm 0.001)}
\\
\midrule
Rehearsal methods   & 
\begin{tabular}{@{}l@{}}
ER \\ PL \\Exp \\ BIR \\ A-GEM 
\end{tabular} &
\begin{tabular}{@{}c@{}}
0.185 \small{(\textpm 0.001)} 
\\ \textbf{0.211 \small{(\textpm 0.002)}}  
\\ 0.185 \small{(\textpm 0.005)}
\\ 0.058 \small{(\textpm 0.002)} 
\\ 0.08 \small{(\textpm 0.012)} 

\end{tabular} &
\begin{tabular}{@{}c@{}}
0.23 \small{(\textpm 0.002)} 
\\  \textbf{0.246 \small{(\textpm 0.002)}} 
\\ 0.223 \small{(\textpm 0.007)}
\\ 0.094 \small{(\textpm 0.009)} 
\\ 0.066 \small{(\textpm 0.011)} 

\end{tabular} &
\begin{tabular}{@{}c@{}}
0.242 \small{(\textpm 0.002)} 
\\  \textbf{0.25 \small{(\textpm 0.009)}} 
\\ 0.222 \small{(\textpm 0.007)}
\\ 0.083 \small{(\textpm 0.007)} 
\\ 0.062 \small{(\textpm 0.003)} 

\end{tabular} &
\begin{tabular}{@{}c@{}}
0.169 \small{(\textpm 0.024)} 
\\  \textbf{0.23 \small{(\textpm 0.004)}} 
\\ 0.206 \small{(\textpm 0.005)}
\\ 0.061 \small{(\textpm 0.005)} 
\\ 0.051 \small{(\textpm 0.001)} 

\end{tabular} &
\begin{tabular}{@{}c@{}}
0.113 \small{(\textpm 0.001)} 
\\  \textbf{0.184 \small{(\textpm 0.004)}} 
\\ 0.161 \small{(\textpm 0.005)}
\\ 0.052 \small{(\textpm 0.002)} 
\\ 0.0506 \small{(\textpm 0.001)} 

\end{tabular}
\\
\midrule
Template Based   & 
 iCaRL
 & 0.09 \small{(\textpm 0.016)}
 & 0.059 \small{(\textpm 0.009)}
 & 0.068 \small{(\textpm 0.014)}
 & 0.059 \small{(\textpm 0.006)}
 & 0.041 \small{(\textpm 0.005)}
\\

\bottomrule
\end{tabular}
\end{small}
\end{center}
\end{table*}

\subsection{Varying the total size of training data for each class (SplitCIFAR-100 and SplitTinyImageNet protocols)}
\begin{table*}[h]
\caption{SplitCIFAR-100: Test accuracy in the final phase as we change the total size of training data for each class. The bracketed values indicate one $SD$ when running experiments with $N = 5$ seeds. LB: Lower Baseline, UB: Upper Baseline, EWC: Elastic Weight Consolidation, SI: Synaptic Intelligence, LwF: Learning without Forgetting, ER: Experience Replay, BIR: Brain-Inspired Replay, A-GEM: Averaged Gradient Episodic Memory, \textit{PL: Power-law}, \textit{Exp: Exponential}, iCaRL: Incremental Classifier and Representation Learning}
\label{cl-comparions-varying-training-data-cifar}
\vskip 0.15in
\begin{center}
\begin{small}
\begin{tabular}{llcccc}
\toprule
Type & Method & \multicolumn{4}{c}{Proportion of training size in SplitCIFAR-100}\\ 
& & 10\% & 20\% & 50\% & 100\% \\
\midrule
Baselines   & 
\begin{tabular}{@{}l@{}}
LB \\ UB
\end{tabular} 
&
\begin{tabular}{@{}c@{}}
0.064\small{(\textpm 0.006)}
\\ 
0.339 \small{(\textpm 0.005)}
\end{tabular}
&
\begin{tabular}{@{}c@{}}
0.069 \small{(\textpm 0.005)}
\\ 
0.39 \small{(\textpm 0.005)}
\end{tabular}
&
\begin{tabular}{@{}c@{}}
0.073 \small{(\textpm 0.006)}
\\ 
0.458 \small{(\textpm 0.006)}
\end{tabular}
&
\begin{tabular}{@{}c@{}}
0.076 \small{(\textpm 0.005)}
\\ 
0.505 \small{(\textpm 0.002)}
\end{tabular}
\\
\midrule
Parameter Regularization   & 
\begin{tabular}{@{}l@{}}
EWC \\ SI
\end{tabular}
& 
\begin{tabular}{@{}c@{}}
0.064 \small{(\textpm 0.006)}\\ 0.069 \small{(\textpm 0.007)}

\end{tabular}
& 
\begin{tabular}{@{}c@{}}
0.069 \small{(\textpm 0.006)} \\ 0.076 \small{(\textpm 0.006)}

\end{tabular}
& 
\begin{tabular}{@{}c@{}}
0.072 \small{(\textpm 0.006)} \\ 0.079 \small{(\textpm 0.004)}
\end{tabular}
& 
\begin{tabular}{@{}c@{}}
0.079 \small{(\textpm 0.007)} \\ 0.081 \small{(\textpm 0.003)}
\end{tabular}
\\
\midrule
Functional Regularization 
& LwF
& 0.198 \small{(\textpm 0.006)}
& 0.245 \small{(\textpm 0.004)}
& 0.264 \small{(\textpm 0.01)}
& 0.255 \small{(\textpm 0.009)}
\\
\midrule
Rehearsal methods   & 
\begin{tabular}{@{}l@{}}
ER \\ PL \\Exp \\ BIR \\ A-GEM 
\end{tabular} &
\begin{tabular}{@{}c@{}}
0.238 \small{(\textpm 0.007)} 
\\  0.268 \small{(\textpm 0.006)}
\\ 0.24 \small{(\textpm 0.067)}
\\  \textbf{0.328 \small{(\textpm 0.009)}}
\\ 0.149 \small{(\textpm 0.014)} 

\end{tabular} &
\begin{tabular}{@{}c@{}}
0.29 \small{(\textpm 0.004)} 
\\  0.322 \small{(\textpm 0.007)}
\\ 0.271 \small{(\textpm 0.008)}
\\ \textbf{0.342 \small{(\textpm 0.004)}}
\\ 0.175 \small{(\textpm 0.022)} 

\end{tabular} &
\begin{tabular}{@{}c@{}}
0.356 \small{(\textpm 0.006)} 
\\  \textbf{0.385 \small{(\textpm 0.004)}} 
\\ 0.331 \small{(\textpm 0.008)}
\\ 0.309 \small{(\textpm 0.011)} 
\\ 0.236 \small{(\textpm 0.052)} 

\end{tabular} &
\begin{tabular}{@{}c@{}}
0.399 \small{(\textpm 0.006)} 
\\  \textbf{0.429 \small{(\textpm 0.006)}} 
\\ 0.373 \small{(\textpm 0.006)}
\\ 0.222 \small{(\textpm 0.011)} 
\\ 0.276 \small{(\textpm 0.012)} 

\end{tabular}
\\
\midrule
Template Based   & 
 iCaRL
 & 0.272 \small{(\textpm 0.009)}
 & 0.312 \small{(\textpm 0.007)}
 & 0.302 \small{(\textpm 0.005)}
 & 0.329 \small{(\textpm 0.01)}
\\

\bottomrule
\end{tabular}
\end{small}
\end{center}
\end{table*}

\begin{table*}[h]
\caption{SplitTinyImagenet: Test accuracy in the final phase as we change the total size of training data for each class. The bracketed values indicate one $SD$ when running experiments with $N = 5$ seeds. LB: Lower Baseline, UB: Upper Baseline, EWC: Elastic Weight Consolidation, SI: Synaptic Intelligence, LwF: Learning without Forgetting, ER: Experience Replay, BIR: Brain-Inspired Replay, A-GEM: Averaged Gradient Episodic Memory, \textit{PL: Power-law}, \textit{Exp: Exponential}, iCaRL: Incremental Classifier and Representation Learning}
\label{cl-comparions-varying-training-data-total-tiny}
\vskip 0.15in
\begin{center}
\begin{small}
\begin{tabular}{llcccc}
\toprule
Type & Method & \multicolumn{4}{c}{Proportion of training size in SplitCIFAR-100}\\ 
& & 10\% & 20\% & 50\% & 100\% \\
\midrule
Baselines   & 
\begin{tabular}{@{}l@{}}
LB \\ UB
\end{tabular} 
&
\begin{tabular}{@{}c@{}}
0.064\small{(\textpm 0.006)}
\\ 
0.187 \small{(\textpm 0.005)}
\end{tabular}
&
\begin{tabular}{@{}c@{}}
0.069 \small{(\textpm 0.005)}
\\ 
0.224 \small{(\textpm 0.005)}
\end{tabular}
&
\begin{tabular}{@{}c@{}}
0.073 \small{(\textpm 0.006)}
\\ 
0.286 \small{(\textpm 0.004)}
\end{tabular}
&
\begin{tabular}{@{}c@{}}
0.076 \small{(\textpm 0.005)}
\\ 
0.341 \small{(\textpm 0.004)}
\end{tabular}
\\
\midrule
Parameter Regularization   & 
\begin{tabular}{@{}l@{}}
EWC \\ SI
\end{tabular}
& 
\begin{tabular}{@{}c@{}}
0.042 \small{(\textpm 0.001)}\\ 0.041 \small{(\textpm 0.002)}

\end{tabular}
& 
\begin{tabular}{@{}c@{}}
0.049 \small{(\textpm 0.006)} \\ 0.046 \small{(\textpm 0.002)}

\end{tabular}
& 
\begin{tabular}{@{}c@{}}
0.053 \small{(\textpm 0.005)} \\ 0.054 \small{(\textpm 0.002)}
\end{tabular}
& 
\begin{tabular}{@{}c@{}}
0.059 \small{(\textpm 0.002)} \\ 0.058 \small{(\textpm 0.001)}
\end{tabular}
\\
\midrule
Functional Regularization 
& LwF
& 0.198 \small{(\textpm 0.006)}
& 0.245 \small{(\textpm 0.004)}
& 0.264 \small{(\textpm 0.01)}
& 0.255 \small{(\textpm 0.009)}
\\
\midrule
Rehearsal methods   & 
\begin{tabular}{@{}l@{}}
ER \\ PL \\Exp \\ BIR \\ A-GEM
\end{tabular} &
\begin{tabular}{@{}c@{}}
0.118 \small{(\textpm 0.003)} 
\\  \textbf{0.125 \small{(\textpm 0.006)}}
\\ 0.107 \small{(\textpm 0.006)}
\\  0.081 \small{(\textpm 0.011)}
\\ 0.046 \small{(\textpm 0.01)} 
\end{tabular} &
\begin{tabular}{@{}c@{}}
0.155 \small{(\textpm 0.004)} 
\\  \textbf{0.169 \small{(\textpm 0.005)}}
\\ 0.146 \small{(\textpm 0.009)}
\\ 0.106 \small{(\textpm 0.013)}
\\ 0.054 \small{(\textpm 0.006)} 

\end{tabular} &
\begin{tabular}{@{}c@{}}
0.201 \small{(\textpm 0.004)} 
\\  \textbf{0.224 \small{(\textpm 0.003)}} 
\\ 0.193 \small{(\textpm 0.008)}
\\ 0.092 \small{(\textpm 0.012)} 
\\ 0.051 \small{(\textpm 0.002)} 

\end{tabular} &
\begin{tabular}{@{}c@{}}
0.242 \small{(\textpm 0.002)} 
\\  \textbf{0.25 \small{(\textpm 0.009)}} 
\\ 0.222 \small{(\textpm 0.007)}
\\ 0.083 \small{(\textpm 0.007)} 
\\ 0.062 \small{(\textpm 0.003)} 

\end{tabular}
\\
\midrule
Template Based   & 
 iCaRL
 & 0.058 \small{(\textpm 0.008)}
 & 0.057 \small{(\textpm 0.006)}
 & 0.062 \small{(\textpm 0.004)}
 & 0.068 \small{(\textpm 0.014)}
\\

\bottomrule
\end{tabular}
\end{small}
\end{center}
\end{table*}

Reducing the number of training images per class increases the complexity of the learning task, as the models have to perform similarly on the test data with a limited training size. We investigate different proportions of class training sizes (10\%, 20\%, 50\%, and 100\% of the total class size). The goal of these experiments is to analyze whether models trained in power-law training setup collapse with less training data, or whether their performance remains comparable to the upper baseline.

Tables \ref{cl-comparions-varying-training-data-cifar} and \ref{cl-comparions-varying-training-data-total-tiny} show the performance of the models with varying training data for each class. The power-law model continues to perform better compared to other techniques at most training sizes. The only exception is BIR, which outperforms the power-law environment for 10\% and 20\% of training data in SplitCIFAR-100. However, with the increase in the training size, the performance of BI-R goes down. The power-law environments' performance is poor at a training data size of 10\%; however, for larger training data sizes the performance is significantly better than other approaches.

\subsection{Description of comparative CI mitigation techniques from prior literature}

\subsubsection{Baselines}
\begin{itemize}
    \item \textbf{Lower Baseline:} Inspired from "lower target" in \citet{van2022three}, each training phase of this training setup only has access to the data for the newly introduced classes, and there is no rehearsal of previous classes. This is akin to a regular neural network architecture with sequential learning of new tasks. We expect this environment to show high levels of CI.

    \item \textbf{Upper Baseline:} In this training setup, the model is trained over data from all classes in the same phase. This is akin to a non-continual learning scenario where the model is trained over all the data at once.
\end{itemize}

\subsubsection{Parameter regularization}

Parameter regularization approaches mitigate CI by penalizing the weight updates that are not orthogonal to the relevant weights of the previous classes. In addition to the classification loss, another loss term also guides the parameter updates: $\mathcal{L}_{\text{total}} = \mathcal{L}_{\text{C}} + \mathcal{L}_{\text{parameter-reg}}$, where $\mathcal{L}_{\text{C}}$ is the classification loss and $\mathcal{L}_{\text{parameter-reg}}$ is the parameter regularization loss.

\begin{itemize}
    \item \textbf{Elastic Weight Consolidation:} EWC \cite{DBLP:journals/corr/KirkpatrickPRVD16} is a regularization-based approach where the model adapts the weight-updating process to limit the shift in weights that are relevant to previous tasks. A quadratic regularization loss is added to the training loss to avoid catastrophic forgetting. For more details, refer to \citet{DBLP:journals/corr/KirkpatrickPRVD16}

    \item \textbf{Synaptic Intelligence:} Instead of having individual loss for each phase, SI \cite{zenke2017continual} consists of a single quadratic loss which is calculated by estimating the parameter importance for the previous K-1 phases. Using a single penalty term reduces the accretion of losses, resulting in poor performance with increasing phases. For more details, refer to \citet{zenke2017continual} 
\end{itemize}

\subsubsection{Functional regularization}

Similar to parameter regularization, functional regularization also adds a penalty term to the training loss to avoid catastrophic interference. However, instead of weighing in parameter importance for previous contexts, these methods prohibit the change of network output on certain anchor points. Training loss in functional regularization: $\mathcal{L}_{\text{total}} = \mathcal{L}_{\text{C}} + \mathcal{L}_{\text{func-reg}}$, where $\mathcal{L}_{\text{C}}$ is the classification loss and $\mathcal{L}_{\text{func-reg}}$ is the functional regularization loss term.

\begin{itemize}
    \item \textbf{Learning without Forgetting:} Our current implementation of LwF is inspired from \cite{van2022three}. LwF performs knowledge distillation by keeping track of model outputs for the same input samples over different phases and computes a combined loss over the ground truth new task output while the old model output is stored in memory. For more details refer to \citet{li2017learning, van2022three}.

\end{itemize}

\subsubsection{Rehearsal-based methods}

Rehearsal-based methods, also known as replay-based methods \cite{deepSLDA, gem, onlineIncrementalLearningVector}, use some learning samples from previous tasks as training samples upon the introduction of a new task. The samples of previous tasks can either be retained in a buffer upon introduction (iCaRL, ER) or generated by a separate generative model (DGR, BI-R).

\begin{itemize}
    \item \textbf{Experience Replay:} A memory buffer stores a fixed amount of randomly chosen instances from each of the previously introduced classes. These act as anchor points for previously seen classes as new classes are introduced.
    \item \textbf{Averaged - Gradient Episodic Memory:} A-GEM \cite{chaudhry2018efficient} differs from ER in the training objective. The algorithm minimizes the loss of the new classes and prohibits the increase in rehearsal loss. Both algorithms use a fixed amount of randomly chosen instances from each class for rehearsal.

    \item \textbf{Brain Inspired - Replay:} BI-R \cite{van2020brain} is a replay-based approach inspired by the replay mechanism in human brains. The replay of hidden representations and features in BI-R is done by a separate generative model which continuously updates over phases. The separate generative model is inspired by and proxies a human hippocampus. 
\end{itemize}

\subsubsection{Template-based classification}

In these approaches, the model extracts features from the input data and learns a representative template which is then used for classification. These templates can be a generative model, a prototype, or a cluster.

\begin{itemize}
    \item \textbf{Incremental Classifier and Representation Learning} iCaRL \cite{icarl} identifies the nearest prototype to the extracted features. It trains a feature extractor network and uses rehearsal with model distillation to ensure that there is no catastrophic forgetting in the feature extractor.  
\end{itemize}


\subsection{ Various data where power-law frequency distribution is observed}

\begin{table*}[!h]
\centering
  \resizebox{\textwidth}{!}{
  
  \begin{tabular}{lc}
    \hline
    Various data where power-law frequency distribution is observed  & References  \\
        
    \hline
    Re-tweets & \cite{cvetojevic2018analyzing,lu2014frequency} \\
    The frequency of occurrence of unique words in a novel (Zipf's Law) & \cite{newman2005power}\\
    Number of distinct interaction partners in protein interaction networks & \cite{ito2000toward}\\
    Degree of known nodes in the internet network representation in autonomous systems & \cite{holme2007radial}\\
    Number of long-distance calls received in the United States in a single day & \cite{abello2002functional, aiello2000random}\\
    Number of unique signers required for sign language recognition (SLR) algorithms to recognize signs accurately& \cite{Bhardwaj_et_al}\\
   Number of examples of a sign required for SLR algorithm to accurately recognize a particular sign& \cite{Bhardwaj_et_al}\\
    Test errors in deep learning algorithms follow a power-law with the number of samples seen per example& \cite{meir2020power}\\
    Number of people affected by electrical blackouts in the United States between 1984 and 2002 & \cite{newman2005power}\\
     Average price of NFTs by categories & \cite{nadini2021mapping}\\
     Degree of items in recommender system graphs & \cite{Milojevi__2010, newman2005power}\\
     Prompts in a text to image models (Diffusion DB) & \cite{Xie_2023}\\
     Phoneme frequencies in collection of illustrative texts & \cite{baird2022blowing}\\
    \hline

\end{tabular}
}
  \caption{Empirical evidence for the natural occurrence of power-law frequency distributions }
  \label{tab:plaw_in_practice}
\end{table*}

\newpage
\subsection{Data distribution in decreasing power-law and exponential curve}

\begin{table*}[h]
\caption{Example distribution of rehearsal data for the classes introduced in each phase in the power-law training environment for 10 phases. Each Group consists of samples from ten classes from the TinyImageNet dataset.}

  \resizebox{0.93\textwidth}{!}{
  \begin{tabular}{lcccccccccccl}
    \hline
   & Group 1 &  Group 2 &  Group 3 &  Group 4 &  Group 5 &  Group 6 &  Group 7 &  Group 8 &  Group 9 &  Group 10 \\
    
    \hline
    
   Phase 1 & 10000 & 0 &     0 & 0 &     0 & 0 &     0 & 0 &     0 &     0 \\
Phase 2 & 5000 & 10000 & 0 &     0 & 0 &     0 & 0 &     0 & 0 &     0 \\
 Phase 3 & 3333 & 5000 & 10000 & 0 &     0 & 0 &     0 & 0 &     0 &     0 \\
 Phase 4 & 2500 &  3333 & 5000 & 10000 & 0 &     0 & 0 &     0 & 0 &     0 \\
 Phase 5 & 2000 &  2500 &  3333 & 5000 & 10000 & 0 &     0 & 0 &     0 &     0 \\
 Phase 6 & 1666 &  2000 &  2500 &  3333 & 5000 & 10000 & 0 &     0 & 0 &     0 \\
 Phase 7 & 1428 &  1666 &  2000 &  2500 &  3333 & 5000 & 10000 & 0 &     0 &     0 \\
  Phase 8 & 1250 &  1428 &  1666 &  2000 &  2500 &  3333 & 5000 & 10000 & 0 &     0 \\
  Phase 9 & 1111 &   1250 &  1428 &  1666 &  2000 &  2500 &  3333 & 5000 & 10000 &     0 \\
  Phase 10 & 1000 &   1111 &   1250 &  1428 &  1666 &  2000 &  2500 &  3333 & 5000 & 10000 \\
    \hline

\end{tabular}
}
  
  \label{tab:  power-law_distribution}
\end{table*}

\begin{table*}[h]
\caption{Example distribution of rehearsal data for the classes introduced in each phase in the exponential training environment for 10 phases. Each Group consists of samples from ten classes from the TinyImageNet dataset.}
  \label{tab: exponential_distribution}
  \resizebox{0.93\textwidth}{!}{
  \begin{tabular}{lcccccccccccl}
    \hline
   & Group 1 &  Group 2 &  Group 3 &  Group 4 &  Group 5 &  Group 6 &  Group 7 &  Group 8 &  Group 9 &  Group 10 \\
    
    \hline
    
   Phase 1 & 10000 & 0 &     0 & 0 &     0 & 0 &     0 & 0 &     0 &     0 \\
Phase 2 & 6239 & 10000 & 0 &     0 & 0 &     0 & 0 &     0 & 0 &     0 \\
 Phase 3 & 3892 & 6239 & 10000 & 0 &     0 & 0 &     0 & 0 &     0 &     0 \\
 Phase 4 & 2428 &  3892 & 6239 & 10000 & 0 &     0 & 0 &     0 & 0 &     0 \\
 Phase 5 & 1515 &  2428 &  3892 & 6239 & 10000 & 0 &     0 & 0 &     0 &     0 \\
 Phase 6 & 945 &  1515 &  2428 &  3892 & 6239 & 10000 & 0 &     0 & 0 &     0 \\
 Phase 7 & 589 &  945 &  1515 &  2428 &  3892 & 6239 & 10000 & 0 &     0 &     0 \\
  Phase 8 & 368 &  589 &  945 &  1515 &  2428 &  3892 & 6239 & 10000 & 0 &     0 \\
  Phase 9 & 229 &   368 &  589 &  945 &  1515 &  2428 &  3892 & 6239 & 10000 &     0 \\
  Phase 10 & 143 &   229 &   368 &  589 &  945 &  1515 &  2428 &  3892 & 6239 & 10000 \\
    \hline

\end{tabular}
}
  
\end{table*}

\newpage

\subsection{Distribution constants for the SplitCIFAR-100 protocol}

\subsubsection{Powerlaw}
\begin{table*}[h]
\caption{  Power-law constants and equations for different minimum number of samples for each previous task at the last phase }
  \label{tab:  power-law_equation}
\centering
  \resizebox{0.6\textwidth}{!}{
  \centering
  \begin{tabular}{lllll}
    \hline
     & \multicolumn{2}{c}{Constants} &  \\
    Min. amount of samples of &   \\
     previous phase classes in a new phase& $a$ & $b$ &   Powerlaw Equation \\
    
    (percentage of the total task size) & \\
    
    \hline
    10\% & 5000 & 1.0 & $5000.x^{-1}$ \\
    5\% & 5000 & 1.301 & $5000.x^{-1.301}$ \\
    2\% & 5000 & 1.699  & $5000.x^{-1.699}$ \\
    1\% & 5000 & 2.0 & $5000.x^{-2}$ \\
    
    \hline

\end{tabular}
}
  
\end{table*}

\subsubsection{Exponential}

\begin{table*}
 \caption{  Exponential constants and equations for various minimum number of samples according to the power-law environment for each previous task at the last phase.}
  \label{tab: exp_equation}
\centering
  \resizebox{0.6\textwidth}{!}{
  \centering
  \begin{tabular}{lllll}
    \hline
     & \multicolumn{2}{c}{Constants} &  \\
    Min. amount of samples (in power-law) of &   \\
     previous phase classes in a new phase& $a$ & $b$ &   Exponential Equation \\
    
    (percentage of the total task size) & \\
    
    \hline
    10\% & 8015.987  & 0.472 & $8015.987e^{-0.472x}$ \\
    5\% & 9587.287 & 0.651 & $9587.287e^{-0.651x}$ \\
    2\% & 12434.041 & 0.911  & $12434.041e^{-0.911x}$ \\
    1\% & 13591.41 & 1.0 & $13591.41e^{-x}$ \\
    
    \hline

\end{tabular}
}
 
\end{table*}

\subsection{Distribution constants for the SplitTinyImageNet protocol}

\subsubsection{Powerlaw}
\begin{table*}[h]
\caption{  Power-law constants and equations for different minimum number of samples for each previous task at the last phase }
  \label{tab:  power-law_equation_tiny}
\centering
  \resizebox{0.6\textwidth}{!}{
  \centering
  \begin{tabular}{lllll}
    \hline
     & \multicolumn{2}{c}{Constants} &  \\
    Min. amount of samples of &   \\
     previous phase classes in a new phase& $a$ & $b$ &   Powerlaw Equation \\
    
    (percentage of the total task size) & \\
    
    \hline
    10\% & 10000 & 1.0 & $10000.x^{-1}$ \\
    5\% & 10000 & 1.301 & $10000.x^{-1.301}$ \\
    2\% & 10000 & 1.699  & $10000.x^{-1.699}$ \\
    1\% & 10000 & 2.0 & $10000.x^{-2}$ \\
    
    \hline

\end{tabular}
}
  
\end{table*}

\subsubsection{Exponential}

\begin{table*}[h]
 \caption{  Exponential constants and equations for various minimum number of samples according to the power-law environment for each previous task at the last phase.}
  \label{tab: exp_equation_tiny}
\centering
  \resizebox{0.6\textwidth}{!}{
  \centering
  \begin{tabular}{lllll}
    \hline
     & \multicolumn{2}{c}{Constants} &  \\
    Min. amount of samples (in power-law) of &   \\
     previous phase classes in a new phase& $a$ & $b$ &   Exponential Equation \\
    
    (percentage of the total task size) & \\
    
    \hline
    10\% & 16031.974  & 0.472 & $16031.974e^{-0.472x}$ \\
    5\% & 19174.573 & 0.651 & $19174.573e^{-0.651x}$ \\
    2\% & 24868.081 & 0.911  & $24868.081e^{-0.911x}$ \\
    1\% & 27182.82 & 1.0 & $27182.82e^{-x}$ \\
    
    \hline

\end{tabular}
}
 
\end{table*}
\newpage
\subsection{Model architectures }
\begin{table}[h]
\caption{Details of the model architecture. The CNN layers are pre-trained with on the CIFAR 10 classification task. The pre-trained CNNs have the following properties (channels: [16, 32, 64, 128, 256], kernel size: 3*3, padding:1, stride:1).  MLP: Multilayer Perceptron, P-CNNs: pre-trained Convolutional Neural Networks, BCE: Binary cross-entropy.}
      \label{tab:architecture}
\centering
  \resizebox{0.7\textwidth}{!}{
    
  \begin{tabular}{llll}
    \hline
    \textbf{Property} & \textbf{SplitMNIST} & \textbf{SplitCIFAR-100} & \textbf{SplitTinyImageNet}\\
    
    \hline
    Architecture type & MLP & P-CNNs + MLP & P-CNNs + MLP\\
    Pre-trained Layers & No & Yes & Yes\\
    Number of hidden layers & 2 MLP & 5 P-CNNs + 2 MLP & 5 P-CNNs + 2 MLP \\
    Dropout probability & 0 & 0 & 0\\
    Learning rate & 0.001 & 0.0001 & 0.0001\\
    Batch size & 128 & 256 & 256\\
    Maximum training iterations per task & 2000 & 5000 & 5000\\
    Optimizer & Adam & Adam & Adam\\
    Loss function & BCE & BCE & BCE\\
    Early stopping & No & No & No\\
  \hline
\end{tabular}
}
\end{table}

\newpage
\section{Phase-wise CI mitigation in different techniques }

\subsection{SplitCIFAR-100}

\begin{table*}[h]
    
\caption{SplitCIFAR-100: Average test accuracy for all classes introduced up to phase (n). The bracketed values indicate one $SD$ when running experiments with $N = 5$ seeds. LB: Lower Baseline, UB: Upper Baseline, EWC: Elastic Weight Consolidation, SI: Synaptic Intelligence, LwF: Learning without Forgetting, ER: Experience Replay, BIR: Brain-Inspired Replay, A-GEM: Averaged Gradient Episodic Memory, \textit{PL: Power-law}, \textit{Exp: Exponential}, iCaRL: Incremental Classifier and Representation Learning}
\label{tab:cifar-cl-phase-all-task}
\vskip 0.15in
\begin{center}
\begin{small}
\begin{tabular}{llccccc}
\toprule
Type & Method & \multicolumn{5}{c}{Phase}\\ 
& & 2 & 4 & 6 & 8 & 10\\
\midrule
Baselines   & 
\begin{tabular}{@{}l@{}}
LB \\ UB
\end{tabular} &
\begin{tabular}{@{}c@{}}
0.396 \small{(\textpm 0.014)}
\\ 
-
\end{tabular}
&
\begin{tabular}{@{}c@{}}
0.192\small{(\textpm 0.334)}
\\ 
-
\end{tabular}
&
\begin{tabular}{@{}c@{}}
0.132 \small{(\textpm 0.004)}
\\ 
-
\end{tabular}
&
\begin{tabular}{@{}c@{}}
0.1 \small{(\textpm 0.002)}
\\ 
-
\end{tabular}
&
\begin{tabular}{@{}c@{}}
0.076 \small{(\textpm 0.005)}
\\ 
0.505 \small{(\textpm 0.002)}
\end{tabular}
\\
\midrule
Parameter Regularization   & 
\begin{tabular}{@{}l@{}}
EWC \\ SI
\end{tabular} &
\begin{tabular}{@{}c@{}}
0.394 \small{(\textpm 0.013)} \\ 0.393 \small{(\textpm 0.014)}

\end{tabular} 
& 
\begin{tabular}{@{}c@{}}
0.192 \small{(\textpm 0.008)} \\ 0.204 \small{(\textpm 0.011)}

\end{tabular}
& 
\begin{tabular}{@{}c@{}}
0.132 \small{(\textpm 0.003)} \\ 0.143 \small{(\textpm 0.005)}

\end{tabular}
& 
\begin{tabular}{@{}c@{}}
0.1 \small{(\textpm 0.002)} \\ 0.109 \small{(\textpm 0.006)}

\end{tabular}
& 
\begin{tabular}{@{}c@{}}
0.079 \small{(\textpm 0.007)} \\ 0.081 \small{(\textpm 0.003)}

\end{tabular}
\\
\midrule
Functional Regularization 
& LwF
& 0.577 \small{(\textpm 0.037)}
& 0.457 \small{(\textpm 0.018)}
& 0.387 \small{(\textpm 0.009)}
& 0.312 \small{(\textpm 0.012)}
& 0.255 \small{(\textpm 0.009)}
\\
\midrule
Rehearsal methods   & 
\begin{tabular}{@{}l@{}}
ER \\ BIR \\ A-GEM \\ PL \\EXP 
\end{tabular} &
\begin{tabular}{@{}c@{}}
0.677 \small{(\textpm 0.051)} 
\\ 0.664 \small{(\textpm 0.061)} 
\\ 0.428 \small{(\textpm 0.033)} 
\\ 0.702 \small{(\textpm 0.046)}
\\ 0.706 \small{(\textpm 0.046)}
\end{tabular} &
\begin{tabular}{@{}c@{}}
0.56 \small{(\textpm 0.028)} 
\\ 0.49 \small{(\textpm 0.031)} 
\\ 0.273 \small{(\textpm 0.024)} 
\\  0.589 \small{(\textpm 0.021)} 
\\ 0.595 \small{(\textpm 0.024)}
\end{tabular} &
\begin{tabular}{@{}c@{}}
0.498 \small{(\textpm 0.017)} 
\\ 0.392 \small{(\textpm 0.014)} 
\\ 0.276 \small{(\textpm 0.015)} 
\\  0.526 \small{(\textpm 0.017)} 
\\ 0.515 \small{(\textpm 0.013)}
\end{tabular} &
\begin{tabular}{@{}c@{}}
0.444 \small{(\textpm 0.007)} 
\\ 0.305 \small{(\textpm 0.005)} 
\\ 0.217 \small{(\textpm 0.009)} 
\\  0.471 \small{(\textpm 0.007)} 
\\ 0.44 \small{(\textpm 0.008)}
\end{tabular} &
\begin{tabular}{@{}c@{}}
0.399 \small{(\textpm 0.006)} 
\\ 0.253 \small{(\textpm 0.012)} 
\\ 0.214 \small{(\textpm 0.046)} 
\\  0.429 \small{(\textpm 0.006)}
\\ 0.373 \small{(\textpm 0.006)}
\end{tabular}
\\
\midrule
Template Based   & 
 iCaRL
 & 0.527 \small{(\textpm 0.265)}
 & 0.449 \small{(\textpm 0.22)}
 & 0.395 \small{(\textpm 0.195)}
 & 0.333 \small{(\textpm 0.162)}
 & 0.329 \small{(\textpm 0.01)}
\\

\bottomrule
\end{tabular}
\end{small}
\end{center}
\end{table*}

\begin{table*}[h]
    
\caption{SplitCIFAR-100: Average test accuracy for \textbf{\textit{previously seen}} classes up to phase (n). The bracketed values indicate one $SD$ when running experiments with $N = 5$ seeds. LB: Lower Baseline, UB: Upper Baseline, EWC: Elastic Weight Consolidation, SI: Synaptic Intelligence, LwF: Learning without Forgetting, ER: Experience Replay, BIR: Brain-Inspired Replay, A-GEM: Averaged Gradient Episodic Memory, \textit{PL: Power-law}, \textit{Exp: Exponential}, iCaRL: Incremental Classifier and Representation Learning}
\label{tab:cifar-cl-phase-old-task}
\vskip 0.15in
\begin{center}
\begin{small}
\begin{tabular}{llccccc}
\toprule
Type & Method & \multicolumn{5}{c}{Phase}\\ 
& & 2 & 4 & 6 & 8 & 10\\
\midrule
Baselines   & 
\begin{tabular}{@{}l@{}}
LB \\ UB
\end{tabular} &
\begin{tabular}{@{}c@{}}
0 \small{(\textpm 0)}
\\ 
-
\end{tabular}
&
\begin{tabular}{@{}c@{}}
0\small{(\textpm 0)}
\\ 
-
\end{tabular}
&
\begin{tabular}{@{}c@{}}
0 \small{(\textpm 0)}
\\ 
-
\end{tabular}
&
\begin{tabular}{@{}c@{}}
0 \small{(\textpm 0)}
\\ 
-
\end{tabular}
&
\begin{tabular}{@{}c@{}}
0 \small{(\textpm 0)}
\\ 
-

\end{tabular}
\\
\midrule
Parameter Regularization   & 
\begin{tabular}{@{}l@{}}
EWC \\ SI
\end{tabular} &
\begin{tabular}{@{}c@{}}
0 \small{(\textpm 0)} \\ 0 \small{(\textpm 0)}

\end{tabular} 
& 
\begin{tabular}{@{}c@{}}
0 \small{(\textpm 0)} \\ 0.015 \small{(\textpm 0.008)}

\end{tabular}
& 
\begin{tabular}{@{}c@{}}
0.001 \small{(\textpm 0)} \\ 0.014 \small{(\textpm 0.005)}

\end{tabular}
& 
\begin{tabular}{@{}c@{}}
0.001 \small{(\textpm 0)} \\ 0.012 \small{(\textpm 0.005)}

\end{tabular}
& 
\begin{tabular}{@{}c@{}}
0.005 \small{(\textpm 0.003)} \\ 0.009 \small{(\textpm 0.003)}

\end{tabular}
\\
\midrule
Functional Regularization 
& LwF
& 0.758 \small{(\textpm 0.054)}
& 0.46 \small{(\textpm 0.033)}
& 0.363 \small{(\textpm 0.009)}
& 0.274 \small{(\textpm 0.013)}
& 0.225 \small{(\textpm 0.012)}
\\
\midrule
Rehearsal methods   & 
\begin{tabular}{@{}l@{}}
ER \\ BIR \\ A-GEM \\ PL \\EXP 
\end{tabular} &
\begin{tabular}{@{}c@{}}
0.595 \small{(\textpm 0.077)} 
\\ 0.572 \small{(\textpm 0.1)} 
\\ 0.068 \small{(\textpm 0.04)} 
\\ 0.661 \small{(\textpm 0.066)}
\\ 0.678 \small{(\textpm 0.067)}
\end{tabular} &
\begin{tabular}{@{}c@{}}
0.507 \small{(\textpm 0.04)} 
\\ 0.408 \small{(\textpm 0.044)} 
\\ 0.104 \small{(\textpm 0.04)} 
\\  0.552 \small{(\textpm 0.033)} 
\\ 0.56 \small{(\textpm 0.034)}
\end{tabular} &
\begin{tabular}{@{}c@{}}
0.454 \small{(\textpm 0.016)} 
\\ 0.319 \small{(\textpm 0.011)} 
\\ 0.173 \small{(\textpm 0.016)} 
\\  0.491 \small{(\textpm 0.015)} 
\\ 0.478 \small{(\textpm 0.013)}
\end{tabular} &
\begin{tabular}{@{}c@{}}
0.406 \small{(\textpm 0.01)} 
\\ 0.238 \small{(\textpm 0.009)} 
\\ 0.133 \small{(\textpm 0.01)} 
\\  0.439 \small{(\textpm 0.011)} 
\\ 0.403 \small{(\textpm 0.01)}
\end{tabular} &
\begin{tabular}{@{}c@{}}
0.373 \small{(\textpm 0.005)} 
\\ 0.201 \small{(\textpm 0.013)} 
\\ 0.154 \small{(\textpm 0.057)} 
\\  0.406 \small{(\textpm 0.004)}
\\ 0.343 \small{(\textpm 0.007)}
\end{tabular}
\\
\midrule
Template Based   & 
 iCaRL
 & 0.534 \small{(\textpm 0.271)}
 & 0.437 \small{(\textpm 0.214)}
 & 0.375 \small{(\textpm 0.185)}
 & 0.312 \small{(\textpm 0.152)}
 & 0.306 \small{(\textpm 0.012)}
\\

\bottomrule
\end{tabular}
\end{small}
\end{center}
\end{table*}

\begin{table*}[h]
    
\caption{  SplitCIFAR-100: Average test accuracy for \textbf{\textit{newly introduced}} classes up to phase (n). The bracketed values indicate one $SD$ when running experiments with $N = 5$ seeds. LB: Lower Baseline, UB: Upper Baseline, EWC: Elastic Weight Consolidation, SI: Synaptic Intelligence, LwF: Learning without Forgetting, ER: Experience Replay, BIR: Brain-Inspired Replay, A-GEM: Averaged Gradient Episodic Memory, \textit{PL: Power-law}, \textit{Exp: Exponential}, iCaRL: Incremental Classifier and Representation Learning}
\label{tab:cifar-cl-phase-new-task}
\vskip 0.15in
\begin{center}
\begin{small}
\begin{tabular}{llccccc}
\toprule
Type & Method & \multicolumn{5}{c}{Phase}\\ 
& & 2 & 4 & 6 & 8 & 10\\
\midrule
Baselines   & 
\begin{tabular}{@{}l@{}}
LB \\ UB
\end{tabular} &
\begin{tabular}{@{}c@{}}
0.791 \small{(\textpm 0.029)}
\\ 
-
\end{tabular}
&
\begin{tabular}{@{}c@{}}
0.769\small{(\textpm 0.037)}
\\ 
-
\end{tabular}
&
\begin{tabular}{@{}c@{}}
0.795 \small{(\textpm 0.021)}
\\ 
-
\end{tabular}
&
\begin{tabular}{@{}c@{}}
0.803 \small{(\textpm 0.016)}
\\ 
-
\end{tabular}
&
\begin{tabular}{@{}c@{}}
0.762 \small{(\textpm 0.048)}
\\ 
-

\end{tabular}
\\
\midrule
Parameter Regularization   & 
\begin{tabular}{@{}l@{}}
EWC \\ SI
\end{tabular} &
\begin{tabular}{@{}c@{}}
0.787 \small{(\textpm 0.027)} \\ 0.786 \small{(\textpm 0.029)}

\end{tabular} 
& 
\begin{tabular}{@{}c@{}}
0.769 \small{(\textpm 0.032)} \\ 0.77 \small{(\textpm 0.036)}

\end{tabular}
& 
\begin{tabular}{@{}c@{}}
0.785 \small{(\textpm 0.017)} \\ 0.786 \small{(\textpm 0.02)}

\end{tabular}
& 
\begin{tabular}{@{}c@{}}
0.793 \small{(\textpm 0.016)} \\ 0.79 \small{(\textpm 0.022)}

\end{tabular}
& 
\begin{tabular}{@{}c@{}}
0.751 \small{(\textpm 0.056)} \\ 0.732 \small{(\textpm 0.052)}

\end{tabular}
\\
\midrule
Functional Regularization 
& LwF
& 0.397 \small{(\textpm 0.026)}
& 0.45 \small{(\textpm 0.043)}
& 0.507 \small{(\textpm 0.032)}
& 0.572 \small{(\textpm 0.035)}
& 0.517 \small{(\textpm 0.066)}
\\
\midrule
Rehearsal methods   & 
\begin{tabular}{@{}l@{}}
ER \\ BIR \\ A-GEM \\ PL \\Exp 
\end{tabular} &
\begin{tabular}{@{}c@{}}
0.759 \small{(\textpm 0.026)} 
\\ 0.756 \small{(\textpm 0.025)} 
\\ 0.788 \small{(\textpm 0.026)} 
\\ 0.743 \small{(\textpm 0.027)}
\\ 0.735 \small{(\textpm 0.027)}
\end{tabular} &
\begin{tabular}{@{}c@{}}
0.721 \small{(\textpm 0.031)} 
\\ 0.735 \small{(\textpm 0.034)} 
\\ 0.778 \small{(\textpm 0.032)} 
\\  0.703 \small{(\textpm 0.032)} 
\\ 0.697 \small{(\textpm 0.038)}
\end{tabular} &
\begin{tabular}{@{}c@{}}
0.718 \small{(\textpm 0.031)} 
\\ 0.755 \small{(\textpm 0.027)} 
\\ 0.791 \small{(\textpm 0.019)} 
\\  0.698 \small{(\textpm 0.029)} 
\\ 0.699 \small{(\textpm 0.027)}
\end{tabular} &
\begin{tabular}{@{}c@{}}
0.712 \small{(\textpm 0.023)} 
\\ 0.775 \small{(\textpm 0.019)} 
\\ 0.804 \small{(\textpm 0.013)} 
\\  0.691 \small{(\textpm 0.025)} 
\\ 0.697 \small{(\textpm 0.021)}
\end{tabular} &
\begin{tabular}{@{}c@{}}
0.639 \small{(\textpm 0.054)} 
\\ 0.721 \small{(\textpm 0.057)} 
\\ 0.749 \small{(\textpm 0.058)} 
\\  0.636 \small{(\textpm 0.049)}
\\ 0.65 \small{(\textpm 0.047)}
\end{tabular}
\\
\midrule
Template Based   & 
 iCaRL
 & 0.521 \small{(\textpm 0.259)}
 & 0.485 \small{(\textpm 0.248)}
 & 0.5 \small{(\textpm 0.245)}
 & 0.479 \small{(\textpm 0.24)}
 & 0.539 \small{(\textpm 0.055)}
\\

\bottomrule
\end{tabular}
\end{small}
\end{center}
\end{table*}
\newpage

\subsection{SplitTinyImagenet }

\begin{table*}[h]

\caption{SplitTinyImagenet: Average test accuracy for all classes introduced up to phase (n). The bracketed values indicate one $SD$ when running experiments with $N = 5$ seeds. LB: Lower Baseline, UB: Upper Baseline, EWC: Elastic Weight Consolidation, SI: Synaptic Intelligence, LwF: Learning without Forgetting, ER: Experience Replay, BIR: Brain-Inspired Replay, A-GEM: Averaged Gradient Episodic Memory, \textit{PL: Power-law}, \textit{Exp: Exponential}, iCaRL: Incremental Classifier and Representation Learning}
\label{tab:tiny-cl-phase-all-task}
\vskip 0.15in
\begin{center}
\begin{small}
\begin{tabular}{llccccc}
\toprule
Type & Method & \multicolumn{5}{c}{Phase}\\ 
& & 2 & 4 & 6 & 8 & 10\\
\midrule
Baselines   & 
\begin{tabular}{@{}l@{}}
LB \\ UB
\end{tabular} &
\begin{tabular}{@{}c@{}}
0.299 \small{(\textpm 0.009)}
\\ 
-
\end{tabular}
&
\begin{tabular}{@{}c@{}}
0.163\small{(\textpm 0.002)}
\\ 
-
\end{tabular}
&
\begin{tabular}{@{}c@{}}
0.111 \small{(\textpm 0.002)}
\\ 
-
\end{tabular}
&
\begin{tabular}{@{}c@{}}
0.076 \small{(\textpm 0.002)}
\\ 
-
\end{tabular}
&
\begin{tabular}{@{}c@{}}
0.062 \small{(\textpm 0.002)}
\\ 
0.341 \small{(\textpm 0.004)}
\end{tabular}
\\
\midrule
Parameter Regularization   & 
\begin{tabular}{@{}l@{}}
EWC \\ SI
\end{tabular} &
\begin{tabular}{@{}c@{}}
0.289 \small{(\textpm 0.003)} \\ 0.286 \small{(\textpm 0.006)}

\end{tabular} 
& 
\begin{tabular}{@{}c@{}}
0.159 \small{(\textpm 0.002)} \\ 0.161 \small{(\textpm 0.002)}

\end{tabular}
& 
\begin{tabular}{@{}c@{}}
0.108 \small{(\textpm 0.003)} \\ 0.109 \small{(\textpm 0.002)}

\end{tabular}
& 
\begin{tabular}{@{}c@{}}
0.074 \small{(\textpm 0.001)} \\ 0.073 \small{(\textpm 0.002)}

\end{tabular}
& 
\begin{tabular}{@{}c@{}}
0.059 \small{(\textpm 0.002)} \\ 0.058 \small{(\textpm 0.001)}

\end{tabular}
\\
\midrule
Functional Regularization 
& LwF
& 0.406 \small{(\textpm 0.01)}
& 0.312 \small{(\textpm 0.006)}
& 0.234 \small{(\textpm 0.006)}
& 0.167 \small{(\textpm 0.003)}
& 0.131 \small{(\textpm 0.004)}
\\
\midrule
Rehearsal methods   & 
\begin{tabular}{@{}l@{}}
ER \\ BIR \\ A-GEM \\ PL \\EXP 
\end{tabular} &
\begin{tabular}{@{}c@{}}
0.449 \small{(\textpm 0.011)} 
\\ 0.354 \small{(\textpm 0.043)} 
\\ 0.336 \small{(\textpm 0.016)} 
\\ 0.475 \small{(\textpm 0.015)}
\\ 0.487 \small{(\textpm 0.006)}
\end{tabular} &
\begin{tabular}{@{}c@{}}
0.344 \small{(\textpm 0.006)} 
\\ 0.175 \small{(\textpm 0.012)} 
\\ 0.165 \small{(\textpm 0.009)} 
\\  0.358 \small{(\textpm 0.014)} 
\\ 0.379 \small{(\textpm 0.004)}
\end{tabular} &
\begin{tabular}{@{}c@{}}
0.311 \small{(\textpm 0.005)}  
\\ 0.124 \small{(\textpm 0.008)} 
\\ 0.112 \small{(\textpm 0.002)} 
\\  0.329 \small{(\textpm 0.004)} 
\\ 0.309 \small{(\textpm 0.009)}
\end{tabular} &
\begin{tabular}{@{}c@{}}
0.265 \small{(\textpm 0.006)} 
\\ 0.1 \small{(\textpm 0.008)} 
\\ 0.077 \small{(\textpm 0.003)} 
\\  0.266 \small{(\textpm 0.008)} 
\\ 0.264 \small{(\textpm 0.009)}
\end{tabular} &
\begin{tabular}{@{}c@{}}
0.242 \small{(\textpm 0.002)}
\\ 0.083 \small{(\textpm 0.007)} 
\\ 0.062 \small{(\textpm 0.003)} 
\\  0.25 \small{(\textpm 0.009)}
\\ 0.222 \small{(\textpm 0.007)}
\end{tabular}
\\
\midrule
Template Based   & 
 iCaRL
 & 0.257 \small{(\textpm 0.008)}
 & 0.174 \small{(\textpm 0.005)}
 & 0.143 \small{(\textpm 0.003)}
 & 0.112 \small{(\textpm 0.007)}
 & 0.068 \small{(\textpm 0.014)}
\\

\bottomrule
\end{tabular}
\end{small}
\end{center}
\end{table*}

\begin{table*}[h]
    
\caption{SplitTinyImagenet: Average test accuracy for \textbf{\textit{previously seen}} classes up to phase (n). The bracketed values indicate one $SD$ when running experiments with $N = 5$ seeds. LB: Lower Baseline, UB: Upper Baseline, EWC: Elastic Weight Consolidation, SI: Synaptic Intelligence, LwF: Learning without Forgetting, ER: Experience Replay, BIR: Brain-Inspired Replay, A-GEM: Averaged Gradient Episodic Memory, \textit{PL: Power-law}, \textit{Exp: Exponential}, iCaRL: Incremental Classifier and Representation Learning}
\label{tab:tiny-cl-phase-old-task}
\vskip 0.15in
\begin{center}
\begin{small}
\begin{tabular}{llccccc}
\toprule
Type & Method & \multicolumn{5}{c}{Phase}\\ 
& & 2 & 4 & 6 & 8 & 10\\
\midrule
Baselines   & 
\begin{tabular}{@{}l@{}}
LB \\ UB
\end{tabular} &
\begin{tabular}{@{}c@{}}
0 \small{(\textpm 0)}
\\ 
-
\end{tabular}
&
\begin{tabular}{@{}c@{}}
0\small{(\textpm 0)}
\\ 
-
\end{tabular}
&
\begin{tabular}{@{}c@{}}
0 \small{(\textpm 0)}
\\ 
-
\end{tabular}
&
\begin{tabular}{@{}c@{}}
0 \small{(\textpm 0)}
\\ 
-
\end{tabular}
&
\begin{tabular}{@{}c@{}}
0 \small{(\textpm 0)}
\\ 
-

\end{tabular}
\\
\midrule
Parameter Regularization   & 
\begin{tabular}{@{}l@{}}
EWC \\ SI
\end{tabular} &
\begin{tabular}{@{}c@{}}
0 \small{(\textpm 0)} \\ 0 \small{(\textpm 0)}

\end{tabular} 
& 
\begin{tabular}{@{}c@{}}
0 \small{(\textpm 0)} \\ 0.002 \small{(\textpm 0.002)}

\end{tabular}
& 
\begin{tabular}{@{}c@{}}
0.001 \small{(\textpm 0)} \\ 0.001 \small{(\textpm 0.001)}

\end{tabular}
& 
\begin{tabular}{@{}c@{}}
0.001 \small{(\textpm 0)} \\ 0.001 \small{(\textpm 0.0)}

\end{tabular}
& 
\begin{tabular}{@{}c@{}}
0.001 \small{(\textpm 0.001)} \\ 0.001 \small{(\textpm 0.0)}

\end{tabular}
\\
\midrule
Functional Regularization 
& LwF
& 0.518 \small{(\textpm 0.013)}
& 0.262 \small{(\textpm 0.014)}
& 0.164 \small{(\textpm 0.005)}
& 0.119 \small{(\textpm 0.002)}
& 0.086 \small{(\textpm 0.005)}
\\
\midrule
Rehearsal methods   & 
\begin{tabular}{@{}l@{}}
ER \\ BIR \\ A-GEM \\ PL \\EXP 
\end{tabular} &
\begin{tabular}{@{}c@{}}
0.348 \small{(\textpm 0.011)} 
\\ 0.129 \small{(\textpm 0.037)} 
\\ 0.081 \small{(\textpm 0.035)} 
\\ 0.399 \small{(\textpm 0.022)}
\\ 0.425 \small{(\textpm 0.007)}
\end{tabular} &
\begin{tabular}{@{}c@{}}
0.262 \small{(\textpm 0.008)} 
\\ 0.047 \small{(\textpm 0.008)} 
\\ 0.006 \small{(\textpm 0.008)} 
\\  0.293 \small{(\textpm 0.016)} 
\\ 0.316 \small{(\textpm 0.006)}
\end{tabular} &
\begin{tabular}{@{}c@{}}
0.251 \small{(\textpm 0.007)} 
\\ 0.035 \small{(\textpm 0.01)} 
\\ 0.002 \small{(\textpm 0.001)} 
\\  0.281 \small{(\textpm 0.006)} 
\\ 0.259 \small{(\textpm 0.008)}
\end{tabular} &
\begin{tabular}{@{}c@{}}
0.229 \small{(\textpm 0.006)} 
\\ 0.033 \small{(\textpm 0.006)} 
\\ 0.002 \small{(\textpm 0.002)} 
\\  0.238 \small{(\textpm 0.006)} 
\\ 0.229 \small{(\textpm 0.001)}
\end{tabular} &
\begin{tabular}{@{}c@{}}
0.211 \small{(\textpm 0.002)} 
\\ 0.027 \small{(\textpm 0.088)} 
\\ 0.003 \small{(\textpm 0.003)} 
\\  0.224 \small{(\textpm 0.009)}
\\ 0.192 \small{(\textpm 0.005)}
\end{tabular}
\\
\midrule
Template Based   & 
 iCaRL
 & 0.455 \small{(\textpm 0.014)}
 & 0.206 \small{(\textpm 0.006)}
 & 0.147 \small{(\textpm 0.004)}
 & 0.12 \small{(\textpm 0.007)}
 & 0.07 \small{(\textpm 0.014)}
\\

\bottomrule
\end{tabular}
\end{small}
\end{center}
\end{table*}

\begin{table*}[h]
    
\caption{  SplitTinyImagenet: Average test accuracy for \textbf{\textit{newly introduced}} classes up to phase (n). The bracketed values indicate one $SD$ when running experiments with $N = 5$ seeds. LB: Lower Baseline, UB: Upper Baseline, EWC: Elastic Weight Consolidation, SI: Synaptic Intelligence, LwF: Learning without Forgetting, ER: Experience Replay, BIR: Brain-Inspired Replay, A-GEM: Averaged Gradient Episodic Memory, \textit{PL: Power-law}, \textit{Exp: Exponential}, iCaRL: Incremental Classifier and Representation Learning}
\label{tab:tiny-cl-phase-new-task}
\vskip 0.15in
\begin{center}
\begin{small}
\begin{tabular}{llccccc}
\toprule
Type & Method & \multicolumn{5}{c}{Phase}\\ 
& & 2 & 4 & 6 & 8 & 10\\
\midrule
Baselines   & 
\begin{tabular}{@{}l@{}}
LB \\ UB
\end{tabular} &
\begin{tabular}{@{}c@{}}
0.598 \small{(\textpm 0.019)}
\\ 
-
\end{tabular}
&
\begin{tabular}{@{}c@{}}
0.654\small{(\textpm 0.009)}
\\ 
-
\end{tabular}
&
\begin{tabular}{@{}c@{}}
0.665 \small{(\textpm 0.011)}
\\ 
-
\end{tabular}
&
\begin{tabular}{@{}c@{}}
0.609 \small{(\textpm 0.012)}
\\ 
-
\end{tabular}
&
\begin{tabular}{@{}c@{}}
0.617 \small{(\textpm 0.016)}
\\ 
-

\end{tabular}
\\
\midrule
Parameter Regularization   & 
\begin{tabular}{@{}l@{}}
EWC \\ SI
\end{tabular} &
\begin{tabular}{@{}c@{}}
0.579 \small{(\textpm 0.005)} \\ 0.571 \small{(\textpm 0.012)}

\end{tabular} 
& 
\begin{tabular}{@{}c@{}}
0.637 \small{(\textpm 0.009)} \\ 0.64 \small{(\textpm 0.007)}

\end{tabular}
& 
\begin{tabular}{@{}c@{}}
0.647 \small{(\textpm 0.019)} \\ 0.648 \small{(\textpm 0.017)}

\end{tabular}
& 
\begin{tabular}{@{}c@{}}
0.584 \small{(\textpm 0.004)} \\ 0.575 \small{(\textpm 0.016)}

\end{tabular}
& 
\begin{tabular}{@{}c@{}}
0.577 \small{(\textpm 0.013)} \\ 0.572 \small{(\textpm 0.006)}

\end{tabular}
\\
\midrule
Functional Regularization 
& LwF
& 0.295 \small{(\textpm 0.011)}
& 0.463 \small{(\textpm 0.024)}
& 0.579 \small{(\textpm 0.014)}
& 0.499 \small{(\textpm 0.011)}
& 0.533 \small{(\textpm 0.007)}
\\
\midrule
Rehearsal methods   & 
\begin{tabular}{@{}l@{}}
ER \\ BIR \\ A-GEM \\ PL \\Exp 
\end{tabular} &
\begin{tabular}{@{}c@{}}
0.55 \small{(\textpm 0.015)} 
\\ 0.579 \small{(\textpm 0.05)} 
\\ 0.592 \small{(\textpm 0.016)} 
\\ 0.551 \small{(\textpm 0.01)}
\\ 0.548 \small{(\textpm 0.009)}
\end{tabular} &
\begin{tabular}{@{}c@{}}
0.591 \small{(\textpm 0.01)} 
\\ 0.56 \small{(\textpm 0.023)} 
\\ 0.642 \small{(\textpm 0.017)} 
\\  0.553 \small{(\textpm 0.013)} 
\\ 0.567 \small{(\textpm 0.011)}
\end{tabular} &
\begin{tabular}{@{}c@{}}
0.611 \small{(\textpm 0.015)} 
\\ 0.568 \small{(\textpm 0.021)} 
\\ 0.66 \small{(\textpm 0.01)} 
\\  0.57 \small{(\textpm 0.024)} 
\\ 0.559 \small{(\textpm 0.022)}
\end{tabular} &
\begin{tabular}{@{}c@{}}
0.516 \small{(\textpm 0.01)} 
\\ 0568 \small{(\textpm 0.035)} 
\\ 0.597 \small{(\textpm 0.015)} 
\\  0.462 \small{(\textpm 0.027)} 
\\ 0.505 \small{(\textpm 0.01)}
\end{tabular} &
\begin{tabular}{@{}c@{}}
0.519 \small{(\textpm 0.003)}
\\ 0.586 \small{(\textpm 0.017)} 
\\ 0.594 \small{(\textpm 0.011)} 
\\  0.479 \small{(\textpm 0.014)}
\\ 0.488 \small{(\textpm 0.021)}
\end{tabular}
\\
\midrule
Template Based   & 
 iCaRL
 & 0.059 \small{(\textpm 0.011)}
 & 0.079 \small{(\textpm 0.008)}
 & 0.12 \small{(\textpm 0.008)}
 & 0.054 \small{(\textpm 0.008)}
 & 0.044 \small{(\textpm 0.019)}
\\

\bottomrule
\end{tabular}
\end{small}
\end{center}
\end{table*}

\newpage
\subsection{SplitMNIST}

\begin{table*}[h]
    
\caption{SplitMNIST: Average test accuracy for all classes introduced up to phase ($n$). The bracketed values indicate one $SD$ when running experiments with $N = 5$ seeds. LB: Lower Baseline, UB: Upper Baseline, EWC: Elastic Weight Consolidation, SI: Synaptic Intelligence, LwF: Learning without Forgetting, ER: Experience Replay, BIR: Brain-Inspired Replay, A-GEM: Averaged Gradient Episodic Memory, \textit{PL: Power-law}, \textit{Exp: Exponential}, iCaRL: Incremental Classifier and Representation Learning}
\label{tab:mnist-cl-phase-all-task}
\vskip 0.15in
\begin{center}
\begin{small}
\begin{tabular}{llccccc}
\toprule
Type & Method & \multicolumn{5}{c}{Phase}\\ 
& & 1 & 2 & 3 & 4 & 5\\
\midrule
Baselines   & 
\begin{tabular}{@{}l@{}}
LB \\ UB
\end{tabular} &
\begin{tabular}{@{}c@{}}
0.996 \small{(\textpm 0.003)}
\\ 
 -
\end{tabular}
&
\begin{tabular}{@{}c@{}}
0.497\small{(\textpm 0.001)}
\\ 
 -
\end{tabular}
&
\begin{tabular}{@{}c@{}}
0.332 \small{(\textpm 0.001)}
\\ 
 -
\end{tabular}
&
\begin{tabular}{@{}c@{}}
0.248 \small{(\textpm 0.001)}
\\ 
 -
\end{tabular}
&
\begin{tabular}{@{}c@{}}
0.198 \small{(\textpm 0.001)}
\\ 
0.978 \small{(\textpm 0.0)}
\end{tabular}
\\
\midrule
Parameter Regularization   & 
\begin{tabular}{@{}l@{}}
EWC \\ SI
\end{tabular} &
\begin{tabular}{@{}c@{}}
0.996 \small{(\textpm 0.003)} \\ 0.996 \small{(\textpm 0.003)}
\
\end{tabular} 
& 
\begin{tabular}{@{}c@{}}
0.495 \small{(\textpm 0.003)} \\ 0.495 \small{(\textpm 0.004)}

\end{tabular}
& 
\begin{tabular}{@{}c@{}}
0.331 \small{(\textpm 0.002)} \\ 0.347 \small{(\textpm 0.021)}

\end{tabular}
& 
\begin{tabular}{@{}c@{}}
0.257 \small{(\textpm 0.016)} \\ 0.249 \small{(\textpm 0.004)}

\end{tabular}
& 
\begin{tabular}{@{}c@{}}
0.231 \small{(\textpm 0.046)} \\ 0.208 \small{(\textpm 0.016)}
 
\end{tabular}
\\
\midrule
Functional Regularization 
& LwF
& 0.996 \small{(\textpm 0.003)}
& 0.635 \small{(\textpm 0.052)}
& 0.378 \small{(\textpm 0.012)}
& 0.267 \small{(\textpm 0.013)}
& 0.212 \small{(\textpm 0.005)}
\\
\midrule
Rehearsal methods   & 
\begin{tabular}{@{}l@{}}
ER \\ BIR \\ A-GEM \\ PL \\Exp 
\end{tabular} &
\begin{tabular}{@{}c@{}}
0.996 \small{(\textpm 0.003)} 
\\ 0.997 \small{(\textpm 0.002)} 
\\ 0.996 \small{(\textpm 0.003)} 
\\ 0.996 \small{(\textpm 0.003)} 
\\ 0.996 \small{(\textpm 0.003)}
\end{tabular} &
\begin{tabular}{@{}c@{}}
0.985 \small{(\textpm 0.003)} 
\\ 0.982 \small{(\textpm 0.005)} 
\\ 0.933 \small{(\textpm 0.032)} 
\\  0.987 \small{(\textpm 0.003)} 
\\ 0.988 \small{(\textpm 0.002)}
\end{tabular} &
\begin{tabular}{@{}c@{}}
0.978 \small{(\textpm 0.003)} 
\\ 0.97 \small{(\textpm 0.006)} 
\\ 0.736 \small{(\textpm 0.156)} 
\\  0.975 \small{(\textpm 0.005)} 
\\ 0.977 \small{(\textpm 0.003)}
\end{tabular} &
\begin{tabular}{@{}c@{}}
0.97 \small{(\textpm 0.004)}
\\ 0.957 \small{(\textpm 0.011)} 
\\ 0.691 \small{(\textpm 0.107)} 
\\  0.966 \small{(\textpm 0.007)} 
\\ 0.96 \small{(\textpm 0.011)}
\end{tabular} &
\begin{tabular}{@{}c@{}}
0.965 \small{(\textpm 0.002)} 
\\ 0.944 \small{(\textpm 0.006)} 
\\ 0.7 \small{(\textpm 0.179)} 
\\  0.959 \small{(\textpm 0.003)}
\\ 0.948 \small{(\textpm 0.002)}
\end{tabular}
\\
\midrule
Template Based   & 
 iCaRL
 & 0.527 \small{(\textpm 0.265)}
 & 0.449 \small{(\textpm 0.22)}
 & 0.395 \small{(\textpm 0.195)}
 & 0.333 \small{(\textpm 0.162)}
 & 0.329 \small{(\textpm 0.01)}
\\

\bottomrule
\end{tabular}
\end{small}
\end{center}
\end{table*}

\begin{table*}[h]
    
\caption{SplitMNIST: Average test accuracy for \textbf{\textit{previously seen}} classes up to phase (n). The bracketed values indicate one $SD$ when running experiments with $N = 5$ seeds. LB: Lower Baseline, UB: Upper Baseline, EWC: Elastic Weight Consolidation, SI: Synaptic Intelligence, LwF: Learning without Forgetting, ER: Experience Replay, BIR: Brain-Inspired Replay, A-GEM: Averaged Gradient Episodic Memory, \textit{PL: Power-law}, \textit{Exp: Exponential}, iCaRL: Incremental Classifier and Representation Learning}
\label{tab:mnist-cl-phase-old-task}
\vskip 0.15in
\begin{center}
\begin{small}
\begin{tabular}{llccccc}
\toprule
Type & Method & \multicolumn{5}{c}{Phase}\\ 
& & 1 & 2 & 3 & 4 & 5\\
\midrule
Baselines   & 
\begin{tabular}{@{}l@{}}
LB \\ UB
\end{tabular} &
\begin{tabular}{@{}c@{}}
 - 
\\ 
 -
\end{tabular}
&
\begin{tabular}{@{}c@{}}
0\small{(\textpm 0)}
\\ 
 -
\end{tabular}
&
\begin{tabular}{@{}c@{}}
0 \small{(\textpm 0)}
\\ 
 -
\end{tabular}
&
\begin{tabular}{@{}c@{}}
0 \small{(\textpm 0)}
\\ 
 -
\end{tabular}
&
\begin{tabular}{@{}c@{}}
0 \small{(\textpm 0)}
\\ 
 -
\end{tabular}
\\
\midrule
Parameter Regularization   & 
\begin{tabular}{@{}l@{}}
EWC \\ SI
\end{tabular} &
\begin{tabular}{@{}c@{}}
 - \\  -
\
\end{tabular} 
& 
\begin{tabular}{@{}c@{}}
0 \small{(\textpm 0.001)} \\ 0.001 \small{(\textpm 0.002)}

\end{tabular}
& 
\begin{tabular}{@{}c@{}}
0 \small{(\textpm 0)} \\ 0.027 \small{(\textpm 0.032)}

\end{tabular}
& 
\begin{tabular}{@{}c@{}}
0.013 \small{(\textpm 0.023)} \\ 0.004 \small{(\textpm 0.006)}

\end{tabular}
& 
\begin{tabular}{@{}c@{}}
0.054 \small{(\textpm 0.084)} \\ 0.017 \small{(\textpm 0.027)}
 
\end{tabular}
\\
\midrule
Functional Regularization 
& LwF
&  -
& 0.281 \small{(\textpm 0.102)}
& 0.07 \small{(\textpm 0.018)}
& 0.026 \small{(\textpm 0.017)}
& 0.016 \small{(\textpm 0.005)}
\\
\midrule
Rehearsal methods   & 
\begin{tabular}{@{}l@{}}
ER \\ BIR \\ A-GEM \\ PL \\EXP
\end{tabular} &
\begin{tabular}{@{}c@{}}
 - 
\\  - 
\\  - 
\\  - 
\\  -
\end{tabular} &
\begin{tabular}{@{}c@{}}
0.979 \small{(\textpm 0.004)} 
\\ 0.972 \small{(\textpm 0.008)} 
\\ 0.871 \small{(\textpm 0.062)} 
\\  0.984 \small{(\textpm 0.004)} 
\\ 0.985 \small{(\textpm 0.003)}
\end{tabular} &
\begin{tabular}{@{}c@{}}
0.972 \small{(\textpm 0.006)} 
\\ 0.959 \small{(\textpm 0.009)} 
\\ 0.606 \small{(\textpm 0.236)} 
\\  0.969 \small{(\textpm 0.006)} 
\\ 0.971 \small{(\textpm 0.004)}
\end{tabular} &
\begin{tabular}{@{}c@{}}
0.966 \small{(\textpm 0.005)}
\\ 0.947 \small{(\textpm 0.015)} 
\\ 0.591 \small{(\textpm 0.143)} 
\\  0.96 \small{(\textpm 0.007)} 
\\ 0.952 \small{(\textpm 0.015)}
\end{tabular} &
\begin{tabular}{@{}c@{}}
0.961 \small{(\textpm 0.002)} 
\\ 0.933 \small{(\textpm 0.008)} 
\\ 0.627 \small{(\textpm 0.224)} 
\\  0.952 \small{(\textpm 0.005)}
\\ 0.94 \small{(\textpm 0.003)}
\end{tabular}
\\
\midrule
Template Based   & 
 iCaRL
 & -
 & 0.974 \small{(\textpm 0.02)}
 & 0.966 \small{(\textpm 0.003)}
 & 0.932 \small{(\textpm 0.002)}
 & 0.933 \small{(\textpm 0.01)}
\\

\bottomrule
\end{tabular}
\end{small}
\end{center}
\end{table*}

\begin{table*}[h]
    
\caption{SplitMNIST: Average test accuracy for \textbf{\textit{newly introduced}} classes up to phase (n). The bracketed values indicate one $SD$ when running experiments with $N = 5$ seeds. LB: Lower Baseline, UB: Upper Baseline, EWC: Elastic Weight Consolidation, SI: Synaptic Intelligence, LwF: Learning without Forgetting, ER: Experience Replay, BIR: Brain-Inspired Replay, A-GEM: Averaged Gradient Episodic Memory, \textit{PL: Power-law}, \textit{Exp: Exponential}, iCaRL: Incremental Classifier and Representation Learning}
\label{tab:mnist-cl-phase-new-task}
\vskip 0.15in
\begin{center}
\begin{small}
\begin{tabular}{llccccc}
\toprule
Type & Method & \multicolumn{5}{c}{Phase}\\ 
& & 1 & 2 & 3 & 4 & 5\\
\midrule
Baselines   & 
\begin{tabular}{@{}l@{}}
LB \\ UB
\end{tabular} &
\begin{tabular}{@{}c@{}}
0.996 \small{(\textpm 0.003)}
\\ 
 -
\end{tabular}
&
\begin{tabular}{@{}c@{}}
0.99\small{(\textpm 0.003)}
\\ 
 -
\end{tabular}
&
\begin{tabular}{@{}c@{}}
0.988 \small{(\textpm 0.005)}
\\ 
 -
\end{tabular}
&
\begin{tabular}{@{}c@{}}
0.983 \small{(\textpm 0.005)}
\\ 
 -
\end{tabular}
&
\begin{tabular}{@{}c@{}}
0.984 \small{(\textpm 0.007)}
\\ 
 -
\end{tabular}
\\
\midrule
Parameter Regularization   & 
\begin{tabular}{@{}l@{}}
EWC \\ SI
\end{tabular} &
\begin{tabular}{@{}c@{}}
0.996 \small{(\textpm 0.003)} \\ 0.996 \small{(\textpm 0.003)}
\
\end{tabular} 
& 
\begin{tabular}{@{}c@{}}
0.99 \small{(\textpm 0.006)} \\ 0.988 \small{(\textpm 0.007)}

\end{tabular}
& 
\begin{tabular}{@{}c@{}}
0.994 \small{(\textpm 0.005)} \\ 0.988 \small{(\textpm 0.008)}

\end{tabular}
& 
\begin{tabular}{@{}c@{}}
0.989 \small{(\textpm 0.004)} \\ 0.984 \small{(\textpm 0.005)}

\end{tabular}
& 
\begin{tabular}{@{}c@{}}
0.938 \small{(\textpm 0.108)} \\ 0.973 \small{(\textpm 0.028)}
 
\end{tabular}
\\
\midrule
Functional Regularization 
& LwF
& 0.996 \small{(\textpm 0.003)}
& 0.989 \small{(\textpm 0.004)}
& 0.993 \small{(\textpm 0.005)}
& 0.992 \small{(\textpm 0.002)}
& 0.992 \small{(\textpm 0.006)}
\\
\midrule
Rehearsal methods   & 
\begin{tabular}{@{}l@{}}
ER \\ BIR \\ A-GEM \\ PL \\EXP
\end{tabular} &
\begin{tabular}{@{}c@{}}
0.996 \small{(\textpm 0.003)} 
\\ 0.997 \small{(\textpm 0.002)} 
\\ 0.996 \small{(\textpm 0.003)} 
\\ 0.996 \small{(\textpm 0.003)} 
\\ 0.996 \small{(\textpm 0.003)}
\end{tabular} &
\begin{tabular}{@{}c@{}}
0.991 \small{(\textpm 0.003)} 
\\ 0.993 \small{(\textpm 0.003)} 
\\ 0.994 \small{(\textpm 0.002)} 
\\  0.99 \small{(\textpm 0.003)} 
\\ 0.99 \small{(\textpm 0.003)}
\end{tabular} &
\begin{tabular}{@{}c@{}}
0.99 \small{(\textpm 0.005)} 
\\ 0.991 \small{(\textpm 0.003)} 
\\ 0.996 \small{(\textpm 0.002)} 
\\  0.988 \small{(\textpm 0.005)} 
\\ 0.989 \small{(\textpm 0.005)}
\end{tabular} &
\begin{tabular}{@{}c@{}}
0.983 \small{(\textpm 0.002)}
\\ 0.987 \small{(\textpm 0.004)} 
\\ 0.993 \small{(\textpm 0.002)} 
\\  0.983 \small{(\textpm 0.005)} 
\\ 0.984 \small{(\textpm 0.004)}
\end{tabular} &
\begin{tabular}{@{}c@{}}
0.982 \small{(\textpm 0.006)} 
\\ 0.988 \small{(\textpm 0.005)} 
\\ 0.993 \small{(\textpm 0.004)} 
\\  0.984 \small{(\textpm 0.007)}
\\ 0.983 \small{(\textpm 0.005)}
\end{tabular}
\\
\midrule
Template Based   & 
 iCaRL
 & 0.997 \small{(\textpm 0.001)}
 & 0.993 \small{(\textpm 0.002)}
 & 0.992 \small{(\textpm 0.005)}
 & 0.986 \small{(\textpm 0.007)}
 & 0.983 \small{(\textpm 0.01)}
\\

\bottomrule
\end{tabular}
\end{small}
\end{center}
\end{table*}



\end{document}